\pdfoutput=1
\documentclass[11pt]{article}
\usepackage{color,soul}
\usepackage[most]{tcolorbox}
\usepackage{tabularray}
\usepackage{siunitx}

\usepackage{ACL2023}
\usepackage{adjustbox}
\usepackage{arydshln}
\usepackage{times}
\usepackage{latexsym}
\usepackage{booktabs}
\usepackage{makecell}
\usepackage[T1]{fontenc}

\usepackage[utf8]{inputenc}

\usepackage{microtype}

\usepackage{inconsolata}

\usepackage{amsmath} 
\usepackage{multicol}
\usepackage{float}
\usepackage{bm}
\usepackage{url}
\usepackage{makecell}
\usepackage{enumitem}
\usepackage{framed}
\usepackage{amsfonts}
\usepackage{multirow}
\usepackage{soul}
\definecolor{codegray}{rgb}{0.9,0.9,0.9}
\usepackage{mdframed}
\usepackage{caption}
\usepackage{amssymb}
\usepackage{booktabs}
\NewTableCommand{\mcc}[1][1]{\multicolumn{#1}{c}}
\newcommand{\eg}{\emph{e.g.,}}

\title{Event Causality Is Key to Computational Story Understanding}

\author{Yidan Sun$^{*}$, Qin Chao\thanks{~~Equal Contribution.}~, \,and Boyang Li\\
  School of Computer Science and Engineering,\\ 
  Nanyang Technological University, Singapore \\
  \texttt{\{SUNY0053, CHAO0009, boyang.li\}@ntu.edu.sg} }

\begin{document}
\maketitle

\begin{abstract}

Cognitive science and symbolic AI research suggest that event causality provides vital information for story understanding. However, machine learning systems for story understanding rarely employ event causality, partially due to the lack of methods that reliably identify open-world causal event relations. Leveraging recent progress in large language models, we present the first method for event causality identification that leads to material improvements in computational story understanding. Our technique sets a new state of the art on the COPES dataset \cite{wang-etal-2023-cola} for causal event relation identification. Further, in the downstream story quality evaluation task, the identified causal relations lead to 3.6-16.6\% relative improvement on correlation with human ratings. In the multimodal story video-text alignment task, we attain 4.1-10.9\% increase on Clip Accuracy and 4.2-13.5\% increase on Sentence IoU. The findings indicate substantial untapped potential for event causality in computational story understanding. The codebase is at \url{https://github.com/insundaycathy/Event-Causality-Extraction}.
\end{abstract}

% Substantial improvements were seen on both tasks, showing that: (1) event causalities extracted by ChatGPT is of high quality; (2) Causality is beneficial to story understanding. 
%Recent research has revealed that the Large Language Model (LLM) exhibits emergent abilities such as in-context learning and chain-of-thought reasoning. More and more attention has been paid to its causality detection capabilities. This paper utilizes a large-scale commonsense causal knowledge dataset, GLUCOSE, and through empirical analysis, provides favourable evidence to demonstrate that ChatGPT is capable of identifying causal relationships between various events occurring in a story. Leveraging the commonsense causal knowledge embedded in LLM, we further demonstrate that this capability can be harnessed to enhance story understanding. For example, 1) by prompting LLM to assess causal relationships in a story, we can improve the correlation between LLM scores and human ratings, and 2) through the analysis of logically supplemented causal chains in text, which goes beyond merely considering temporal precedents, we can improve the accuracy of video-text alignment.
% checklist:
% event causality
% use of '
% misspelling 
% informal language
% better than what
% use active voice
% refrain from using they/this, is ambiguous
% 0.59 million to 590k
% \footnote{$^\ast$ Corresponding Author. \textdagger  Equal Contribution.}

\section{Introduction}
\label{sec:intro}
Stories manifest in various forms in modern society, such as myths, fables, gossip, comic books, bedtime rituals for children, and million-dollar theatrical productions. Stories are theorized to play crucial roles in civilization, from building collective identities \cite{bruce_lincoln_myth:1999} to familiarizing readers with social skills \cite{oatley2008}.  Research on computational story generation \cite{guan2020knowledge,ammanabrolu2021automated,xie2022psychology,yang2022re3,hong2023visual,yang-etal-2023-doc} and understanding \cite{du2021learning,xu2022fairytaleqa, andrus2022enhanced,2022tvshowguess,dong2023corrpus} has gained traction in recent years.

%\hl{this whole part has been restructured to include cognitive science experiments, and story generation and our experiment of adding 'causal'}
Converging evidence points to the information value of event causality in story understanding. Cognitive science indicates that humans heavily rely on event causality in story comprehension \citep{fletcher1988causal, graesser200310}, as reflected by experiments on event recall and prediction \citep{trabasso1985causal,keefe1993time}. Intuitively, causal relations affect story understanding and value judgements. Story events at the end of properly linked causal chains may appear believable (\eg{} the deaths of Romeo and Juliet), even though the events may be unusual. Based on the causes of events, we make moral judgments and assign blame. For instance, the revenge of Hamlet is caused by his uncle murdering his father and hence may be considered just. 

The symbolic AI approach to computational story generation also makes extensive use of human-crafted event causal relations \cite{Meehan-TaleSpin-1976, young1994decomposition, li2010offline, porteous2011controlling,soo2016generate}. Anecdotally, merely adding the word ``causal'' to the ChatGPT prompt of \citet{wang2023chatgpt} leads to a 3\% relative boost in story evaluation (\S \ref{sec:story_eval}).
However, event causal relations are rarely utilized by deep learning-based methods for story understanding,
%(though there are a few attempts to use them in story generation \citep{ammanabrolu2021automated,ye2022neural,TattleTale2022,Kelly-AIIDE-2023})
possibly due to the difficulty in identifying event causal relations in an open-world setting.

In this paper, we argue that causal structures --- the story events and the causal relations among them --- offer crucial and operationalizable information for computational story understanding; we further propose an easy-to-use technique for extracting such relations by prompting large language models (LLMs). With few-shot in-context learning, we enable LLMs to reconstruct causal structures from open-domain, free-form story text. 

To verify the validity of the extracted causal structures, we first compare them against human-annotated causal relations of \citet{Mostafazadeh2020GLUCOSEGA} and \citet{wang-etal-2023-cola}, leveraging a diverse range of LLMs. Empirically, the proposed method performs comparably with and sometimes surpasses supervised state-of-the-art baselines.  

However, even if the causal structures are correct, they may not be of value to story understanding. 
To examine the value of the causal structures, we conduct further tests on two downstream tasks: story quality evaluation \cite{guan2021openmeva} and story video-text alignment \cite{dogan2018neural,sun2022synopses}. 
In story quality evaluation, incorporating the extracted event causal structures improves Kendall's tau relatively by 
 6.4\%-15.6\%.  In story video-text alignment, it  improves clip accuracy by 4.1-10.9\% and sentence IoU by 4.2-13.5\%.

In summary, the experimental results indicate that (1) the simple prompting technique we propose can identify story causal relations with high accuracy, and (2) the identified story structures indeed benefit story understanding tasks. Since the identified structures coincide substantially with human-annotated causal relations, we argue the empirical evidence supports the thesis that automatically extracted event causality facilitates computational story understanding. 
Our contributions can be summarized as:
\begin{itemize}
    \item We propose a simple prompt-based technique for identifying event causal structures from free-form stories in diverse domains. 
    
    \item With the proposed technique, we set a new state of the art on the COPES event causality benchmark \cite{wang-etal-2023-cola}. 
    
    \item To our best knowledge, this is the first work to demonstrate the practical benefits of causal story structures in automated story understanding, leading to substantial improvements on two distinct tasks, the text-only story quality evaluation and the multimodal story video-text alignment.
\end{itemize}

The organization of the paper may be somewhat unconventional. After reviewing the background knowledge of causal reasoning and related work in Section 2, we introduce our method in Section 3. Our method is evaluated on three different tasks related to causal relation extraction and story understanding in Sections 4, 5 and 6, respectively. Each of the last three sections presents its own experiment setup, results, and discussion. 

\section{Background and Related Work}
\label{sec:related_work}

% \subsection{\hl{Definitions of Event Causality}}

% \hl{Various definitions of event causality exist. Here, we list the popular definitions without attempting to debate the merits of each. }
% \begin{itemize}
%     \item \hl{\textit{The multi-factorial causation model}: Proposed by the 19th-century hygienist Max Von Pettenkofer }{\cite{oppenheimer2007invited,morabia2007epidemiologic}}, \hl{this model states that Event A causes Event B if, in combination with other factors, Event A is a necessary or sufficient condition for Event B. }
%     \item \hl{\textit{The counterfactual causation model} }\cite{kim1973causes}: \hl{Event A causes Event B, if when Event A does not occur, Event B will not occur. }

%     \item \hl{\textit{The probabilistic causation model} }\cite{reichenbach1991direction}: \hl{ Event A causes Event B, if the occurrence of Event A raises the probability of Event B occurring. }

%     \item \hl{\textit{The interventionist causation model} }\cite{woodward2007interventionist}:  \hl{Event A causes Event B, if there is a possible intervention on Event A that can change Event B.}
% \end{itemize}

\subsection{Causal Reasoning about Events}
Causal reasoning about the effects and counterfactual effects of actions and events is undoubtedly an important tool in modern scientific thinking \cite{SHOHAM1990213} and an integral  area of AI research \cite{Pearl2018-PEATBO}. However, causality appears surprisingly difficult to define. \citet{Pearl2018-PEATBO} suggests that attempts to define causality are ``unproductive'' (Chapter 1) and we should focus on the benefits of causal reasoning instead. Nevertheless, at the behest of the anonymous reviewers, we present two definitions of event causality, both compatible with our work. 

\noindent \textbf{Definition 1.} We say Event A causes Event B if:
\begin{itemize}
    \item (the multi-factorial definition) in combination with other factors, Event A is a necessary or a sufficient condition for Event B \cite{oppenheimer2007invited,morabia2007epidemiologic}, or
    \item (the probabilistic definition) the occurrence of Event A raises the probability of Event B occurring \cite{reichenbach1991direction}.
\end{itemize}

Event causality is often conditioned on a myriad of other factors and may be neither necessary or sufficient by itself. For example, we may say the event  Alice divorcing Bob is caused by the event Bob having an extramarital affair. However, an affair may not end every marriage, and some marriages end for different reasons. Hence, what we take for the cause is neither necessary nor sufficient for the effect. 

To understand the concept, it is perhaps useful to review common categories of event causality. \citet{trabasso1989logical} provide four categories. First, an event physically causes another event, like kicking a ball causing the ball to move. Second, a physical event causes a psychological reaction, like winning a lottery causing joy. Third, a psychological condition may motivate an action, such as the desire for a driver's license causing someone to take the driving test. Fourth, an event may establish conditions for a second event to happen. An example is that organizing a chess tournament causes someone to become the champion. In this paper, we focus on commonsense interpretations of causality, as exact analysis according to the  definitions is usually infeasible (\eg{} probabilities of events in a story world are hard to determine). 

\subsection{Event Causality in Human Story Understanding}
% need to make it clear that cognitive science focus on information processing and that event causal relations is infromation that's need for story understanding
% rephrase graesser's work
% add in adult and children
Cognitive science research indicates event causality offers crucial information in human narrative comprehension. \citet{gernsbacher1997coherence} discover that when two events in a sentence are joined by a causal connective, the second event is better memorized than if the two events are joined by a non-causal connective. Story events with more causal connections are better recalled and are judged by humans as more important \cite{trabasso1985causal,van1996children}. Furthermore, event causality also influences the prediction of future events in a narrative. \citet{keefe1993time} discover that, immediately after reading an event, words related to the possible effects of the event are recognized faster than unrelated words. 

\subsection{Event Causality in Computational Story Generation}
% ammanabrolus's work does not invlove story plans
% add a sentence to state what is done with story plans.
% need to fact check on whether or not there are works for event causality in story understanding -checked can't find any
%
Event causality has been widely used in computational story generation \cite{lebowitz1985story,bae2008use,swartjes2010whose,TattleTale2022,liu2023magic,Kelly-AIIDE-2023}. Early works in symbolic story generation constructed story plans from human-written action templates that stipulate the preconditions and effects of actions \cite{lebowitz1985story,young1994decomposition,bae2008use,riedl2010narrative,li2010offline,brenner2010creating, swartjes2010whose}. A story plan arranges the story events so that the preconditions of later events are fulfilled by the effects of prior events. 

The reliance on human-crafted knowledge limits story planners to narrow domains. Recent works attempt to utilize large language models commonsense to acquire action templates \cite{ye2022neural,TattleTale2022, spiliopoulou2022events, Kelly-AIIDE-2023}. %However, \citet{spiliopoulou2022events} demonstrated that language models cannot reliably model state changes resulting from events.
\citet{ammanabrolu2021automated} attempt to build story graphs with neural networks trained to perform causal relation completion. However, we are not aware of the utility of the extracted templates and story graphs in story understanding tasks. %\citet{sun2018reading} extract story graphs containing temporal and causal relations and attempt to use the story graph in QA tasks. However, \citet{sun2018reading} did not demonstrate the sole effect of causal relations.
Compared to story planning, which sometimes can operate with a known list of people, objects, and actions, story understanding needs to deal with a vast assortment of narratives in open-world settings. As a result, the ability to identify causal structures in arbitrary stories becomes crucial \cite{caselli2021computational}.

\subsection{Commonsense Causal Reasoning}

%%%%% -prev version about ECI -
% Event Causality Identification (ECI) \cite{girju2003automatic} predicts the existence and type of causal relations between two events. Most existing datasets \cite{mirza2014annotating,caselli2017event, hu2017inference, wang2022maven} represent an event with one word and omit detailed event constituents such as participants, instruments, time and locations; the relations are mostly identified from the surface text, rather than commonsense knowledge. The objective of this paper is different as we aim to perform commonsense causality reasoning \cite{kuipers1984commonsense,roemmele2011choice}, which is to identify causal relations between contextualized events using commonsense knowledge.

% GLUCOSE \cite{Mostafazadeh2020GLUCOSEGA} and COPES \cite{wang-etal-2023-cola} are two human-annotated event causality detection datasets that are well suited for our purpose. These papers use supervised learning, whereas we apply few-shot in-context learning to leverage the rich knowledge from LLMs. 
%%%%% -prev version-

%(why we need context)% 
The objective of Commonsense Causal Reasoning (CCR) is to identify commonsense causal relations between events from text \cite{kuipers1984commonsense,roemmele2011choice, zhang2022rock}, which is distinct from causal relation identification that require domain expertise, such as medical knowledge \cite{gurulingappa2012development}. Such causal relations are often heavily dependent on context, such as participants, time, and locations of events. For example, the two events ``cooking at home'' and ``cooking at a restaurant'' likely have different causes. The former is likely caused by hunger for food, but the latter is likely caused by the job requirement. COPA \cite{roemmele2011choice}, GLUCOSE \cite{Mostafazadeh2020GLUCOSEGA}, and COPES \cite{wang-etal-2023-cola} are prototypical datasets for CCR. In this paper, we focus on GLUCOSE and COPES, as their problem formulations contain more story context than COPA. 

%Using a question-and-answer format, i.e., COPA \cite{roemmele2011choice}, can bypass simple superficial clues. But as suggested by \cite{kavumba2019choosing,mingyue2021doing}, the superior performance of supervised models on COPA may be attributed to token distribution asymmetry or semantic bias among choices, rather than relying on commonsense knowledge. Meanwhile, studies are pointing out that existing CCR datasets overlook the influence of context on causal relationships, as highlighted by GLUCOSE \cite{Mostafazadeh2020GLUCOSEGA}, ROCK \cite{zhang2022rock} and COLA \cite{wang-etal-2023-cola}, causality can indeed change based on contextual variations. Therefore, as proposed by COLA, finding the cause of an event within a specific context in the free-form text is a crucial metric for assessing a model's ability to detect causality.

Prior works test LLMs on COPA \cite{wei2021finetuned,anil2023palm, gao2023chatgpt}, and LMs on  GLUCOSE and COPES \cite{li2022cis2, colon2023adversarial, wang2023contextualized, wang-etal-2023-cola}. %(T5 and BART and ParaCOMET) on GLUCOSE, their settings are different: cis2:mask the sentence, colon:give hint about the desired output, wang: on good quality subset and only test 5 dimensions), COLA has no LLM tests, only qualitative results reported in their appendix%
%\hl{cite papers that test on COPA, GLUCOSE, and COPES. - finished searching, plz check} 
%\hl{double check our difference.} 
%\citet{sun2018reading} extract temporal and causal relations in stories and apply the extract relations directly to story QA tasks, but does not demonstrate the sole effect of causal relations. 
To the best of our knowledge, this work is the first to quantitatively explore the ability of ChatGPT 3.5 to understand Contextualized CCR (\emph{i.e.,} GLUCOSE and COPES) and its impact on downstream tasks.

% However, the event causality detection models proposed in \citep{Mostafazadeh2020GLUCOSEGA} and \citep{wang-etal-2023-cola} are fine-tuned on everyday children's stories from the ROCStories \cite{mostafazadeh2016corpus} and may not generalize to other stories.

% In this work, we take advantage of the commonsense knowledge embedded in LLMs and propose a prompt for few-shot event causality extraction from free-form story text with LLMs. 

\subsection{Open-ended Generated Story Evaluation}
A critical ingredient in story generation research is the automatic evaluation of story quality, as human comparisons can be expensive and difficult to replicate. Metrics that involve direct comparisons against gold references, such as BLEU \cite{papineni2002bleu}, have limited applicability as there is no single correct story for each writing prompt. Researchers have proposed supervisedly trained techniques \citep{ghazarian2019better, sellam2020bleurt, guan2020union} and in-context learning methods based on LLMs \cite{wang2023chatgpt, chiang2023can, shen2023large}. In this work, we demonstrate that providing event causality information in the LLM context can further enhance the correlation between LLM ratings and human ratings.

\section{Methodology}
\label{sec:event_graph_prompt}

Our objective is to acquire a \emph{causal graph}, a directed graph that contains events as nodes and causal relations as directed edges.
At the core of our approach is an LLM prompt that includes an instruction and a list of story events, as shown in Figure \ref{fig:prompt}. The prompt can contain a number of examples, though we show only one due to length considerations. The prompt requests the LLM to detect and output causal relations among the events. For simplicity, we consider each sentence in the story as an event. 

The output format for the causal relations is \texttt{Edge: (Node A -> Node B)}. In preliminary experiments, we find that the arrow ``\texttt{->}'' notation, similar to the influential DOT language~\cite{dot-language-2015} of graph representation, tends to yield better results than other notations we tried.  

% Using this prompt, the LLM will extract causal relationships between events from the story. The input is a list of nodes, each node represents an event in the story. The extracted relationships are then listed in the form of \footnote{ We believe there might be different ways to express event causality.}, where each node represents an event/sentence and each edge represents a casual relationship between two events. We consider these edges as a representation of the event graph. 
% With this prompt, the LLM can extract all event causality within a story by generating an event graph for each story. In the event graph, each node represents an event and each edge represents a casual relationship between two events.

% This prompt is suitable for extracting event causality with LLM because of the event graphs format and the representation of event causality as arrows (``->''). We notice that without a given output form, LLMs tend to represent event causality as arrows. Perhaps the arrow representation of event causality is well presented in the LLMs' train data. 

\begin{figure}[t]
\begin{tcolorbox}
\small
\tt
\noindent Here is a list of nodes (events) from a story event graph. We want you to fill in the edges of the event graph with causal connections between nodes. An event graph contains nodes and edges. Each node represents an event, and each edge represents the causal connection between two events. \\
\\
Example Input: \\
Node 0: When Dan goes to school in the morning, he has to take the bus. \\
Node 1: One day Dan was running late, and missed the bus to school. \\
Node 2: Dan called his friend Pete, and asked for a ride to school. \\
Node 3: Pete gave Dan a ride to school, but Dan was late for his first class. \\
Node 4: Luckily Dan wasn't late for any of his other classes that day. \\
Example Output: \\
Edge 0: (Node 0 -> Node 1) \\
Edge 1: (Node 1 -> Node 2) \\
Edge 2: (Node 2 -> Node 3) \\
Edge 3: (Node 1 -> Node 3) \\
Edge 4: (Node 3 -> Node 4) \\
\emph{(continue with another five demonstrations)}\\

Now, it is your turn to construct the event graph for the following event list.\\
Event List:\\
Node 0: <S1>\\
Node 1: <S2>\\
Node 2: <S3>\\
Node 3: <S4>\\
Node 4: <S5>\\
Output:
%\textcolor[RGB]{101,42,150}{\texttt{Output:}}
\end{tcolorbox}

\caption{The LLM prompt for event causal relationship extraction.}
\label{fig:prompt}
\end{figure}

\section{Event Causality Extraction}
\label{sec:expt_glucose}
To assess the quality of LLM-extracted event causal relations, we compare against two human-annotated benchmarks: COPES \cite{wang-etal-2023-cola} and GLUCOSE \cite{Mostafazadeh2020GLUCOSEGA}. It is worth noting that the purpose of these experiments is not to seek state-of-the-art performance, but to verify the identified causal relations are of decent quality. However, we still manage to beat state-of-the-art baselines on COPES. 

\paragraph{Task Definition and Datasets} For the COPES task, the input is a pair of events and the output is whether or not a causal relationship exists between the events. COPES contains 340 stories and 1360 event pairs from ROCStories, split 50/50 into the validation set and the test set.

The GLUCOSE dataset contains a number of causal dimensions. We select only Dimensions 1 and 6, which concern causality between events. Given a story and one of its events, the task is to identify all of the direct causes or effects of the event from the story. GLUCOSE paraphrases the identified causes and effects as well as the current event in a subject-verb-object format and applies reference-based evaluation such as BLEU.  
As our technique only outputs causal relations and does not perform paraphrasing, we directly use the original sentences from the story.  

% The GLUCOSE dataset contains 590K human-annotated causal statements from the ROCStories dataset categorized into 10 dimensions. Among these dimensions, we report on dimensions 1 and 6 which pertain to causal relationships between two events (results on other dimensions are in Appendix \ref{app:glucose_dims}).  

% For both benchmarks, the input is a list of events. For GLUCOSE, the output is a list of causal statements in the form of ``\textit{Event A} >Causes/Enables> \textit{Event B}". In the GLUCOSE annotations, \textit{Event A} / \textit{Event B } are simplifications of the sentence in the story that describes event A or event B. In our method, we directly use the original story sentence \hl{why? differing from norm.}. For COPES, the output is a list of all the event pairs in the story and whether or not a causal relationship exists between each event pair. 

%These relationships need to be presented in the form of  and some degree of simplification or paraphrasing is allowed. 

\paragraph{Model}
% For event causality extraction, we experiment with three HuggingFace official implementation LLMs: Llama2-13B-chat-hf\footnote{\label{note1}\url{https://huggingface.co/meta-llama/Llama-2-13b-chat-hf}; \url{https://huggingface.co/tiiuae/falcon-40b-instruct}; \url{https://huggingface.co/01-ai/Yi-34B}}, Falcon-instruction-40B\footnotemark[1\ref{note1}], Yi-34B-chat\footnotemark[1\ref{note1}], and ChatGPT (gpt3.5-turbo).
For event causality extraction, we experiment with four advanced LLMs: Llama2-13B-chat \cite{touvron2023llama}, Falcon-instruction-40B\footnote{\url{https://falconllm.tii.ae/falcon.html}}, Yi-34B-chat\footnote{\url{https://01.ai/}} and ChatGPT-3.5-turbo-0631 \cite{ouyang2022training}.

% \paragraph{Dataset}e
 %\hl{We use the training samples from GLUCOSE and the validation samples from COPES to create the demonstration samples used in our prompt.}

%Note, GLUCOSE requires the model to generate a causal statement that match the human annotation. How and the human written causal statement contains paraphrases and noise. With COPES, the model only does binary classification of whether or not a causal relationship exists. Therefore, COPES is less affected by subjective human annotation. However, since GLUCOSE is larger and more well-known, we report performance on both. 

\paragraph{Evaluation}
For COPES, we follow \citet{wang-etal-2023-cola} and report accuracy, Micro F1, and Macro F1. We use 6 randomly selected stories from the validation set as in-prompt examples and report performance on the COPES test set.

For GLUCOSE, we choose in-prompt examples randomly from the training set and  conduct evaluation on the GLUCOSE test set of 293 stories. As of evaluation metrics, our main objective is to evaluate if the model can accurately distinguish between causally related, positive event pairs and unrelated, negative event pairs. Hence, we compute the precision and recall of the predicted positive class, and combine them as the F1 score. For completeness, 
we also adopt BLEU\footnote{Implementation from \citet{post2018call}} following \citet{Mostafazadeh2020GLUCOSEGA} as well as the BERTscore \citep{zhang2019bertscore} and Sentence BERT similarity \citep{reimers2019sentence}, but these metrics mostly evaluate the surface form and are not as important as F1.

%using SacreBLEU \citep{post2018call}, a standard BLEU implementation calculating the overlap of up to 4-grams between references and predictions.
%with equal weights up to 4-grams at the corpus level on the 3-reference test set.
%However, the human-annotated references of GLUCOSE involve simplifications and paraphrases of the story text which BLEU cannot capture. Therefore, we also measure BERT score \citep{zhang2019bertscore} and Sentence BERT similarity \citep{reimers2019sentence}, which are better for paraphrases. Furthermore, we use F1 score to measure the models' ability to identify the presence of causal relationships. %Due to space limitations, we report the results aggravated over the 10 dimensions and the aggravated over dimensions 1 and 6, results for separated dimensions can be found in Appendix \ref{app:glucose_results_dims}.

\paragraph{Baselines}
For COPES, we directly compare our model with COLA \cite{wang-etal-2023-cola}. Also, we compare against ROCK \cite{zhang2022rock} and ClozePromptScore \cite{tamborrino2020pre} as replicated and reported by \citet{wang-etal-2023-cola}. 

For GLUCOSE, we replicated the two models used by \citet{Mostafazadeh2020GLUCOSEGA} as baselines. We train the same networks (T5 and GPT2-large) on the training split. However, the training set of \citet{Mostafazadeh2020GLUCOSEGA} contains only positive examples, or pairs of causally related events. In order to handle causally unrelated, negative event pairs, which are abundant in open-world stories, we exhaustively add all negative event pairs to the training set, yielding 590K training samples. After that, from pretrained LLM weights, we train baseline network to output a description of the cause or effect for positive cases or ``Nil'' for negative cases. See Appendix \ref{app:finetune_glucose} for more details.

\begin{table}
\centering
\small
\begin{adjustbox} {width=0.45\textwidth}

\begin{tabular}{@{}lccc@{}}
  \toprule
   & Acc. & Micro F1 & Macro F1   \\
  \midrule
  \multicolumn{4}{c}{\emph{Supervised}}\\
  $\text{ClozePromptScore}$ & 62.06 & 45.57 &   58.22 \\ 
 $\text{ROCK}$ & 66.47 & 51.90 &   63.08 \\ 
 $\text{COLA}$ & 70.29 & 57.38 &   67.29 \\
\midrule
 \multicolumn{4}{c}{\emph{Few-shot (Ours)}}\\
 $\text{Falcon-40B-instuct}$ & 65.74 & 41.60 &   58.68 \\ 
 $\text{Llama-2-13B-chat}$ & 71.47 & 47.58 &   63.99 \\ 
 $\text{Yi-34B-chat}$ & 72.94 & 55.98 & 68.22 \\
 $\text{ChatGPT-3.5}$ & \textbf{74.26} & \textbf{57.42} &   \textbf{69.49} \\
\bottomrule
\end{tabular}
\end{adjustbox}
\caption{Performance on COPES. }
\label{tab:copes}
\end{table}

\begin{table}
\centering
\small
\begin{adjustbox} {width=0.45\textwidth}
\begin{tabular}{@{}lcccc@{}}
\toprule
  & F1 & BLEU  & BERTScore & \makecell[c]{BERT \\ Similarity.} \\
\midrule
   \multicolumn{5}{c}{\emph{Supervised}}\\
  $\text{GPT-2}_{\text{large}}$ & 59.54 & 28.92 & 79.86 & 84.64  \\
  $\text{T5}_{\text{large}}$  & \textbf{61.50} & \textbf{31.75} & \textbf{84.34} & \textbf{88.77}  \\
\midrule
  \multicolumn{5}{c}{\emph{Few-Shot (Ours)}}\\
  $\text{Falcon}$ & 28.57 & 13.43 & 38.65 & 25.68  \\
  $\text{Llama-2}$ & 51.70 & 19.77 & 58.22 & 54.82  \\
   %$\text{Llama}_{\text{ensemble}}$ & 22.39 & 67.23 & 66.05 & 57.09 \\ 
  $\text{Yi}$ & 57.95 & 18.95 & 77.42 & 84.32  \\
  $\text{ChatGPT}$ & 60.75 & 21.20 & 75.33 & 80.89  \\
 
\bottomrule
\end{tabular}
\end{adjustbox}
\caption{The BLEU, BERTScore, BERT Similarity, and F1 score on GLUCOSE dataset, averaged over dimensions 1 \& 6. }
\label{tab:glucose_bleu_16}
\end{table}

%,as they are recognized as robust baselines by \citet{wang-etal-2023-cola}. Due to space limitation, we only test the BERT-large as the backbone for these baselines, as it demonstrated superior performance in \citet{wang-etal-2023-cola}.
%For COPES, we directly use the COLA model proposed in their paper as a baseline. We also compare against the ROCK \cite{zhang2022rock} and ClozePromptScore \cite{tamborrino2020pre}, which are strong baselines used in \citet{wang-etal-2023-cola}. Due to space limitations, we use BERT-large as the backbone for the baselines, which yielded the best performance in \citet{wang-etal-2023-cola}.

%Predicting all causality within a story in one sitting requires long input and output sequences that render finetuning difficult. Therefore, we modify the input of the supervised models. Instead of providing a single output statement for the entire story at once, we configure it to address one sentence and one dimension at a time. Hence, the input is a tuple that contains a story, a story sentence, and a dimension; the output is Nil when a causality of the given dimension cannot be found for the given sentence, or a causal statement in the form of `` A >relationship> B'' where either A or B is a paraphrase of the current sentence and ``>relationship>'' describes a causal relationship. Examples of the input and output is shown in Appendix \ref{app:finetune_glucose}. Note that this will set a simplified condition for the supervised baselines that puts ChatGPT at a disadvantage.

\paragraph{Results}

%In Table \ref{tab:glucose_bleu_10}, we compare the few-shot results from LLMs against the supervised baselines. With only 6 examples, $\text{ChatGPT}_{\text{ensemble}}$ can outperform the supervised GPT2 and match supervised T5 trained on 590k causal statements on most metrics. 
In Table \ref{tab:copes}, we show performance on COPES. With only 6 example stories, our technique with ChatGPT outperforms the state-of-the-art (SOTA) supervised model, COLA, by 4.2\% in accuracy and 2.3\% in Macro F1. Furthermore, our method is robust and can generalize to different LLMs. We observe our technique with Yi-34B-chat also outperforms the SOTA on accuracy and Macro-F1 and Llama-2-13B outperforms ClozePromptScore.

Table \ref{tab:glucose_bleu_16} shows performance on GLUCOSE. In terms of F1 score, our technique with ChatGPT-3.5 outperforms supervised GPT2-large model trained on 590K causal statements and performs comparably to the supervised T5 model. Among the opensource models, Yi-34B-chat performs best and can match the GPT2-large baseline on most metrics. On reference-based metrics, LLMs score lower than supervised models mainly because supervised models are trained to imitate paraphrased annotations of GLUCOSE whereas while the few-shot LLMs are not trained or prompted to paraphrase. Hence, the reference-based metrics underestimate the actual performance of our technique. However, on BERT similarity, our few-shot technique with Yi-34B-chat closes the gap with GPT-2.

\paragraph{Case Study}

We show an example of a generated event graph in Figure \ref{fig:glucose_example_1}. More examples are in Appendix \ref{app:case_study}. Here, all edges identified agree with GLUCOSE annotations. The man lying down (Node 0) enables the cat to jump on his stomach (Node 1). This is the fourth category of \citet{trabasso1989logical}. The causality between Node 1 (the cat jumping on the man's stomach) and 3 (the man petting the cat) is twofold. First, cat puts itself within the reach of the man, thereby enabling petting. Also, the cat shows affection for the man, motivating him to reciprocate. Between Node 3 and Node 4 (the cat took a nap), the reciprocated affection comforts the cat and motivates it to stay put, possibly taking a nap. The model outputs two repeated edges (Edge 0 and Edge 3), which can be easily filtered out.

\begin{figure}[t]
\begin{tcolorbox}[enhanced]
\small
\tt
Event List:\\
Node 0: The man laid down for a nap.\\
Node 1: His cat jumped on his stomach.\\
Node 2: That woke the man up.\\
Node 3: The man petted the cat.\\
Node 4: The cat took a nap with the man.\\
Output: \\
Edge 0: (Node 1 -> Node 2) \textcolor{teal}{[Correct]} \\
Edge 1: (Node 0 -> Node 1) \textcolor{teal}{[Correct]} \\
Edge 2: (Node 0 -> Node 2) \textcolor{teal}{[Correct]} \\
Edge 3: (Node 1 -> Node 2) \textcolor{teal}{[Correct]} \\
Edge 4: (Node 1 -> Node 3) \textcolor{teal}{[Correct]} \\
Edge 5: (Node 3 -> Node 4) \textcolor{teal}{[Correct]} \\
\end{tcolorbox}
\caption{Example of ChatGPT-generated event graph. The example is from GLUCOSE, the \textcolor{teal}{[Correct]} labels are not part of the model output.}
\label{fig:glucose_example_1}
\end{figure}

\section{Open-domain Story Evaluation}
\label{sec:story_eval}
%\hl{minor changes: I think we shouldn't say LLM for section 5 as we only use chatgpt, correlation etc. etc. are details}

Having established that the event causality that we identified are quite accurate, we still need to verify if the event causality provide valuable information to actual story understanding tasks. To this end, we conduct two tests. This section describes the first test, automatic story quality evaluation. We generate the quality ratings from ChatGPT with and without the automatically extracted event causal relations, and compare how they correlate with human ratings. 

% Our experiments focus on providing an overall score for machine-generated text based on the quality.

%\noindent  In this subsection, our experiments focus on providing an overall score based on the quality of the story. We will explore how to enhance the correlation between LLM and human ratings in the open-ended generated story by providing LLM with event causality information.
\paragraph{Approach: Quality Ratings Conditioned on Causal Graphs}

We propose a two-stage prompting method that scores the quality of a story conditioned on its causal graph. First, we prompt ChatGPT to generate the causal graph of the story, using the same prompt in Figure \ref{fig:prompt}. Then, we include the causal graph, which contains a list of causal relations between event descriptions, in a scoring prompt that asks ChatGPT to generate an overall score for story quality. The scoring prompt is derived from \citet{wang2023chatgpt} (see Appendix \ref{app:story_eval}). %We test both the zero-shot and the few-shot settings.
%, where a star of 5 indicates excellent grammar and coherence in the story 

We test three settings, zero-shot, in-domain few-shot, and cross-domain few-shot, which differ in examples used in the scoring prompt. In zero-shot, we do not include any examples in the scoring prompt. In in-domain few-shot, we include two example stories we wrote manually in the style of OpenMEVA-ROC and OpenMEVA-WP respectively. In cross-domain few-shot, we include two example stories from OpenMEVA-ROC when testing on OpenMEVA-WP and vice versa. 
In all settings, the causal graph generation stage uses the same six story examples from GLUCOSE.

% Interestingly, in our exploratory experiments, simply adding the term "causal" to the original prompt resulted in a relative improvement of 3\% in dataset-level correlation on ROC dataset. 

%We test the performance of LLM in both the zero-shot setting and when providing an additional high-quality story and a low-quality story as examples.

%\hl{the formatting here looks weird}\\
% \vspace{0.5mm}\\
% \textsc{stage 1: }\textls[50]{generate event graph with the prompt shown in Figure \ref{fig:prompt}}

% \noindent\textsc{stage 2: }\textls[50]{scoring}

% \begin{figure}[t]
%  \centering
%  \includegraphics[width=\linewidth]{stages.pdf}

%  \caption{Prompt for stage 2 of story evaluation.}
% \label{fig:prompt_evalution}
% \end{figure}
% \hl{I think we should change these to figures so that they are easiler to refer to -Agree! thx for the figure}
% \begin{tcolorbox}[breakable, enhanced]
% \small
% \tt
% \raggedright\noindent ... (the same scoring prompt used by \citet{wang2023chatgpt})\\
% We also provide causal connections analyzed by experts, where each event is represented as a node, and the causal connections between these nodes are listed.\\
% Event graph: \\
% \textcolor[RGB]{101,42,150}{\texttt{[list the event graph generated with ChatGPT]}}
% \\
% Your score should reward stories with rich causal chains and penalize those that lack or have confusing causal chains.
% \end{tcolorbox}

\paragraph{Dataset}
% Existing benchmarks are mainly divided into two categories: overall evaluation \citep{ghazarian2019better,guan2020union, guan2021openmeva} and dimension-specific evaluation \citep{chhun2022human, ke2022ctrleval,xie2023next}. 
% This paper focuses on methods for overall scoring. 
The dataset we use is OpenMEVA \citep{guan2021openmeva}. The dataset was acquired
from five different story generation models trained on ROCStories \cite{mostafazadeh2016corpus} and another five trained on WritingPrompt (WP) \cite{fan2018hierarchical}. Each model generates stories from 200 writing prompts. As a result, OpenMEVA consists of two parts: OpenMEVA-ROC and OpenMEVA-WP, each containing 1,000 generated stories. Each story is evaluated by five human annotators, each assigning a score between 1 and 5. The final score for the story is the average of the five.

% Each data point contains a machine-generated story and a human-evaluated score.
%\hl{I still think this is details that do not pertain to the later experiments, and long}

OpenMEVA-ROC and OpenMEVA-WP have different characteristics. Stories in OpenMEVA-ROC always contain 5 sentences and, on average, 34 words. The lengths of stories in OpenMEVA-WP are more varied, with an average of 20 sentences  and 194 words. Qualitatively, we observe that OpenMEVA-ROC mostly retain the style of ROCStories, where the sentences are simple and describe clear-cut events. In comparison, OpenMEVA-WP is much more diverse, containing much non-event content such as conversations and monologues, and much more vague event boundaries. We show two examples in Figure \ref{fig:open-meva-example} of the Appendix. 

%the ROC dataset consists of stories composed of five sentences, while the WP dataset has variable story lengths, with an average of 20 sentences per story. 
%\hl{the notation $rho$ never comes up again}
\paragraph{Evaluation} We employ correlation to assess the similarity between ChatGPT scores and human ratings. We report three well-established correlations: Pearson correlation \citep{benesty2009pearson}, Spearman rank correlation \citep{zar2005spearman}, and Kendall's tau coefficient \cite{kendall1938new}. All of these measures have values ranging from -1 to 1, with values closer to 1 indicating a stronger positive correlation.

We calculate the correlation between human ratings and ChatGPT ratings at two aggregation levels: (1) dataset level, where we measure the correlation between two scoring systems across the entire dataset, and (2) writing prompt level, where we compute the correlation between the two scoring systems for the five stories generated for each writing prompt and then average the results. The formula can be found in Appendix \ref{app:story_eval}.

\begin{table*}[t]
\resizebox{\textwidth}{!}{
\begin{tabular}{@{}llll|lll|lll|lll!{}}
% \begin{tabular}{llll|lll|lll|lll}
\toprule
\multicolumn{1}{c}{\multirow{3}{*}{Metrics}} & \multicolumn{6}{c|}{OpenMEVA-ROC (n=1000)} & \multicolumn{6}{c}{OpenMEVA-WP (n=1000)} \\
\multicolumn{1}{c}{} & \multicolumn{3}{c|}{Writing Prompt Level} & \multicolumn{3}{c|}{Dataset Level} & \multicolumn{3}{c|}{Writing Prompt Level} & \multicolumn{3}{c}{Dataset Level} \\
\cline{2-13}
\multicolumn{1}{c}{} & \multicolumn{1}{c}{Pear.} & \multicolumn{1}{c}{Spear.} & \multicolumn{1}{c|}{Kend.} & \multicolumn{1}{c}{Pear.} & \multicolumn{1}{c}{Spear.} & \multicolumn{1}{c|}{Kend.} & \multicolumn{1}{c}{Pear.} & \multicolumn{1}{c}{Spear.} & \multicolumn{1}{c|}{Kend.} & \multicolumn{1}{c}{Pear.} & \multicolumn{1}{c}{Spear.} & \multicolumn{1}{c}{Kend.} \\ 
\midrule

BART+CNN+Para &0.050 & 0.064& 0.062& 0.062& 0.074& 0.043 & 0.014 & 0.046 & 0.045 & 0.083& 0.077& 0.053 \\
BERTScore-F1 & 0.144 & 0.131 & 0.103 & 0.127 & 0.113 & 0.079 & 0.089 & 0.085 & 0.077 & 0.033 & 0.031 & 0.022 \\
BLEURT $\text{in-domain}^{\ast}$ & - & - & - & 0.316 & - & - & - & - & - & 0.212 & - & - \\
Perplexity & 0.330 & 0.324 & 0.265 & 0.255 & 0.306 & 0.213 & 0.373 & \textbf{0.381} & 0.318 & 0.303 & 0.324 & 0.225 \\
UNION $\text{in-domain}^{\ast}$ & - & - & - & 0.412 & - & - & - & - & - & 0.326 & - & - \\
UNION $\text{cross-domain}^{\ast}$& - & - & - & 0.213 & - & - & -  &  - &  - & 0.229 & - & - \\
Original \text{\citet{wang2023chatgpt}} & 0.490 &0.472 & 0.427 & 0.439 & 0.415 & 0.342 & - & - & -& - & - & -\\ 
\midrule
\multicolumn{3}{l}{\qquad\emph{ChatGPT zero-shot}}& \multicolumn{1}{c|}{}& \multicolumn{3}{c|}{}&\multicolumn{3}{c|}{}&\multicolumn{3}{c}{}\\
Repl. $\text{\citet{wang2023chatgpt}}^{\spadesuit}$ & 0.526 & 0.520 & 0.472 & 0.446 & 0.436 & 0.366 & 0.281  & 0.257 & 0.236 & 0.203 & 0.199 & 0.165 \\
ChatGPT-``causal'' & 0.531 & 0.522 & 0.474 & 0.460 & 0.451 & 0.379 & 0.301  & 0.275 & 0.246 & 0.215  & 0.215 & 0.183 \\
ChatGPT-causal-graph & 0.576 & 0.562 & 0.510 & 0.520 & 0.505 & 0.423 & 0.331 & 0.299 & 0.273 & 0.277 & 0.277 & 0.230 \\
\midrule
\multicolumn{3}{l}{\qquad\emph{ChatGPT in-domain few-shot} }& \multicolumn{1}{c|}{}& \multicolumn{3}{c|}{}&\multicolumn{3}{c|}{}&\multicolumn{3}{c}{}\\
Repl. $\text{\citet{wang2023chatgpt}}^{\spadesuit}$ & 0.553 & 0.526 & 0.466 & 0.498 & 0.496 & 0.398 & 0.313 & 0.291 & 0.257 & 0.269 & 0.262 & 0.208 \\
ChatGPT-``causal'' & 0.560 & 0.537 & 0.480 & 0.501 & 0.503 & 0.402 & 0.327 & 0.305 & 0.270 & 0.276 & 0.276 & 0.218 \\
ChatGPT-causal-graph & \textbf{0.592} & \textbf{0.575} & \textbf{0.520} & \textbf{0.526} & \textbf{0.514} & \textbf{0.425} & 0.339 & 0.313 & 0.285 & 0.284 & 0.294 & 0.246\\
\midrule
\multicolumn{3}{l}{\qquad\emph{ChatGPT cross-domain few-shot}}& \multicolumn{1}{c|}{}& \multicolumn{3}{c|}{}&\multicolumn{3}{c|}{}&\multicolumn{3}{c}{}\\
Repl. $\text{\citet{wang2023chatgpt}}^{\spadesuit}$ & 0.519 & 0.500 & 0.433 & 0.462 & 0.452 & 0.348 & 0.373 & 0.345 & 0.301 & 0.293 & 0.297 & 0.228 \\
ChatGPT-``causal'' & 0.506&	0.513&	0.449&	0.461&	0.463&	0.357&	\textbf{0.404}&	0.353&	0.303&	0.314& 0.316 &	0.243 \\
ChatGPT-causal-graph & 0.547 & 0.530 & 0.459 & 0.498 & 0.482 & 0.370 & 0.387 & 0.367 & \textbf{0.323} & \textbf{0.328} & \textbf{0.328} & \textbf{0.258} \\ 
\bottomrule
\multicolumn{13}{l}{\normalsize {$\ast$: Results are taken from OpenMEVA benchmark \citet{guan2021openmeva} }}\\
\multicolumn{13}{l}{\normalsize {$\spadesuit$: For fair comparison, We replicate \citet{wang2023chatgpt} using the same few-shot settings and ChatGPT model (\texttt{gpt3.5-turbo-}}}\\
\multicolumn{13}{l}{\normalsize {\texttt{0613, temp=0}) as in other experiments.}}\\
\end{tabular}

}
\caption{Writing prompt-level and dataset-level correlations on OpenMEVA.  (Spear.: Spearman correlation; Pear.: Pearson correlation; Kend.: Kendall's Tau).}\label{tab:story_eval}
\end{table*}

\paragraph{Baselines} We compare our method with seven baselines, including two reference-based methods, one hybrid method, two reference-free methods, and two LLM methods. First, the reference-based baselines rate the computer-generated stories by matching them against a human-written story for the same writing prompt. The two baselines are (1) BERTScore \citep{zhang2019bertscore} and 
(2) BARTScore+CNN+Para \citep{yuan2021bartscore}, which computes the perplexity of the text conditioned on the reference text. The hybrid method is (3) BLEURT \citep{sellam2020bleurt}, a neural network that \citet{guan2020union} adapt to evaluate machine stories against a reference. 

The two reference-free baselines are (4) perplexity on GPT-2 \citep{radford2019language}, which gives higher rankings to stories with lower perplexity and (5) UNION \citep{guan2020union}, a neural network trained to discriminate machine stories from human stories; machine stories that are more similar to human stories are considered better. 

In addition, we compare against the ChatGPT prompt of \citet{wang2023chatgpt}, which rates the OpenMEVA-ROC dataset on a scale of 1-5 stars. For fair comparisons, we also replicate this baseline using the same few-shot settings and the same ChatGPT-3.5 model. Finally, we create another variation (ChatGPT-``causal'') by adding the word ``causal'' to the prompt of \citet{wang2023chatgpt}. Details can be found in Appendix \ref{app:story_eval}. 

% UNION is a BERT model fine-tuned\footnote{Note that UNION used ROCStories or WP to train separate models. Specifically, if the models are both trained and evaluated on ROCStories, it is in-domain evaluation. If the models are trained on WP and evaluated on ROCStories, it is a cross-domain evaluation.} to discriminate model-generated texts from human-written ones. 

% And the last one is the LLM-based method: (6) 
% Under the hypothesis that human-written stories are high-quality stories, the more it is close to human-written the story is better. 

\paragraph{Results and Discussion} 
We show results in Table \ref{tab:story_eval}, where our approach is denoted as ChatGPT-causal-graph. 
We observe that event causality provides significant improvements, especially in zero-shot settings. On zero-shot ROC, our method achieves relative improvements of 8.05\% to 16.59\% over the best baseline. On few-shot ROC, our method achieves relative improvements of 3.65\% to 11.59\% over the best baseline.
On zero-shot WP and few-shot WP, causality graph brings even greater gains over the \cite{wang2023chatgpt} baseline, due to the low baseline performance. However, on WP stories, we surpass all baselines only in the cross-domain few-shot condition. 

The performance on WP warrants further analysis. Our technique performs worse with in-domain examples than cross-domain examples, which seems to contradict machine learning commonsense. We attribute this to two reasons. First, the WP stories are longer and hence more difficult to understand as in-context examples. Second, WP stories are less event-centric and have more vague event boundaries than ROC. This may cause errors in ChatGPT-extracted causal graphs, which hurt in-context learning. Instead, when we use the more correct causal graphs from ROC as examples in the prompt, performance improves. 

Though the WP results suggest that our technique for causal graph extraction may not work equally well in all story domains, we emphasize that this is the first work that has ever shown causal graphs provide benefits for \emph{any} computational story understanding task.

\begin{figure*}[t]
 \centering
 \includegraphics[width=0.93\linewidth]{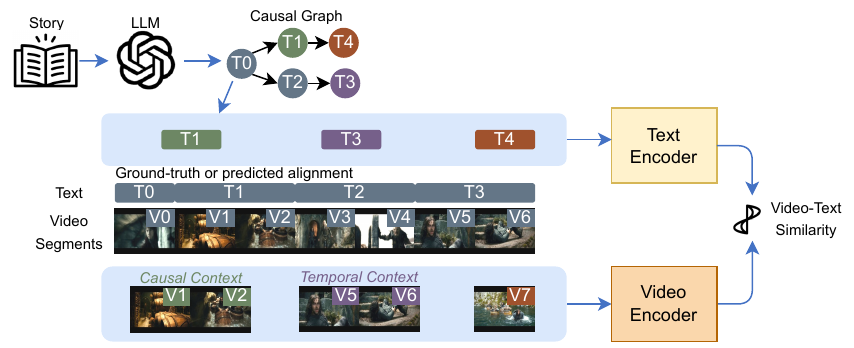}
 \caption{The process of video context identification. The causal context is marked in green, the temporal context  in purple, and the current item in red. For illustration purposes, the number of causal and temporal context items are both set to 2. }
 %\hl{Videos should be numbered V1, V2. T1, T3, and T4 on the top should have equal length. The nodes in the causal graph should have the same color. That is, the colors should have unique meanings. Purple = temporal context. Green = causal context. ``Video-text alignment'' should be ``Ground-truth or predicted alignment''.}}
\label{fig:context-identification}
\end{figure*}

\section{Story Video-text Alignment}
\label{sec:story_alignment}
The second test task for the automatically extracted event causal relations is story video-text alignment. Due to the wide use of storytelling techniques, aligning the video and the text modalities requires significant story understanding \cite{sun2022synopses}. We show that preceding events on the causal graph provide crucial context for this task. Considering the domain gaps between this multimodal task and previous pure text tasks, this experiment supports the argument that our proposed technique can cope with a broad range of real-world stories. 

\paragraph{Task Definition and Datasets} 
The task starts with a movie summary video from YouTube, which contains shots selected from a movie and human narration of the main plotline. The video has been segmented into a number of clips, and the narration has been transcribed into text and segmented into chunks. However, due to modality differences, the text chunks and video clips at the same time may not match each other semantically. The task is weakly supervised; we need to find the correct semantic alignment without training on gold alignment labels. 

We adopt the Synopsis of Movie Narratives (SyMoN) dataset \cite{sun2022synopses} for training, which contains 5,193 movie summary videos. The test set comes from the YouTube Movie Summary (YMS) dataset \cite{dogan2018neural}, which has gold alignment labels. We report results on two different splits and textual chunking levels.

% . YMS collects movie summary videos from YouTube and annotate with sentence-level and sub-sentence-level video correspondence. %\hl{split them into both sentence-level and sub-sentence-level video-text pairs. We report performance at both annotation granularities.} 
 % We report performance on both annotation levels.

\paragraph{Approach: Context-aware video-text alignment}
To align a video sequence and a text sequence, we follow a three-step procedure. First, we encode each video clip and each text chunk together with their temporal and causal contexts. Then, we calculate the cosine similarity between each video-text pair. Finally, we calculate the overall sequence alignment from the individual similarity scores using Dynamic Time Warping (DTW), a sequence alignment algorithm detailed in Appendix \ref{app:alignment}.  

We finetune pretrained visual and textual encoders from UniVL \cite{luo2020univl}. The visual encoder contains an S3D network that encodes 1-second clips into tokens, followed by a Transformer. The text encoder is a Transformer. We denote the encoded features for the $i^{\text{th}}$ text chunk as $\bm t_{i}$ and the encoded features of the $i^{\text{th}}$ video clip as $\bm v_{i}$. The feature vectors are normalized to unit length, so that cosine similarity is simply dot product. With randomly sampled negative text features $\bm t_{k}$ and video features $\bm v_{k}$, we finetune the encoders by minimizing the contrastive NCELoss
\citep{gutmann2010noise}, 
\begin{equation}
\begin{split}
{L}_{\text{NCE}} = & \frac{1}{N} \sum\limits_{i=1}^{N} - \bm{v}_{i}^\top \bm t_{i} 
	+ \log \Big( \exp \bm{v}_{i}^\top\bm t_{i} + \\ & \sum\limits_{k\ne i}^{K}{\exp \bm{v}_{i}^\top\bm t_{k}} + \sum\limits_{k\ne i}^{K}{\exp \bm{v}_{k}^\top\bm t_{i}} \Big),
\end{split}
\end{equation}
where $N$ is the total number of training samples and $K$ is the number of negative samples. 

Note that the training set of \citet{sun2022synopses} does not contain human-annotated alignment. Therefore, we adopt a weakly supervised approach that considers video clips and text chunks with similar temporal positions as positive pairs and randomly sample negative pairs during training. 

When encoding the current video and text, the visual and textual Transformer networks also take in a number of contextual items (video clips or text chunks). These contextual item values are retrieved from a memory bank, which stores the item values from all layers of the Transformer network. After retrieval, the item values are fed to the corresponding Transformer encoder layers. We will discuss how the context is constructed momentarily. 

% At inference time, we are given a video sequence and a text sequence and need to find their alignment. We first compute the pairwise similarity of all videos and text chunks. After that, we compute a global alignment with Dynamic Time Wrapping (DTW),  

% Context is incorporated into the model through the self-attention layer. Specifically, a memory bank stores the self-attention keys and values from the video segments/text that precede that current segment/text. We store a set of keys and values for each visual and text encoder layer, therefore, 24 sets of keys and values are stored in the memory bank. At each time step, we identify the context of the current video segment/text using the causal graph. Then, we retrieve the keys and values of the context video segments/text from the memory bank and append them with the keys and values of the current video segment/text to form the memory-ware key and value. The self-attention score is calculated with the original query, the memory-aware key, and the memory-aware value.

\paragraph{Context Identification}
Here we distinguish between two types of contexts, temporal context and causal context. The temporal context contains $m$ number of items (video clips or textual chunks) that immediately precede the current item being encoded. The causal context contains $c$ items preceding the current item on the causal graph extracted from the entire story text by ChatGPT. 

The causal context for a textual chuck can be directly identified from the causal graph, but the causal context for a video clip requires some extra processing. During training, we first find the textual causal predecessors from the causal graph. Next, the video clips temporally closest to the text predecessor chunks are deemed as the causal context for the current video clip. During inference, we align the two sequences incrementally from the beginning. Given a text predecessor, we use the aligned portion of the two sequences to locate the corresponding video clips as the causal context. The process is illustrated in Figure \ref{fig:context-identification}.

\paragraph{Evaluation Metrics}
Following \citet{dogan2018neural}, we use two evaluation metrics: Clip Accuracy, defined as the temporal proportion of correctly aligned video segments, and Sentence IoU, defined as the intersection-over-union between the aligned video durations and the ground-truth durations.

\paragraph{Baselines}
Our main baseline is the temporal context only, ablated version of our technique, which uses $c+m$ temporal context items instead of $c$ causal context items and $m$ temporal context items. Additionally, we compare against (1) the Minimal Distance and Dynamic Time Warping baseline in NeuMATCH \cite{dogan2018neural}, and (2) the Minimal Distance baseline in SyMoN \cite{sun2022synopses}. Both baselines did not utilize context.

\paragraph{Results and Discussion}
As shown in Table \ref{tab:alignment_result}, incorporating causal context from the identified causal graph yields improvements across the board. The highest improvement for Clip Accuracy is 10.9\% and the highest improvement for Sentence IoU is 13.5\%. Note the SyMoN test split contains 65 videos whereas the NeuMATCH test split contains only 15 videos. Hence, the SyMoN split numbers may be more trustworthy. In the sentence-level SyMoN split, which is arguably more natural than the sub-sentence level, adding event causality improves Clip Accuracy by 7.7\% and Sentence IoU by 8.0\%. These results convincingly demonstrate that the automatically extracted causal graphs provide real benefits in the story video- text alignment task, even though the multimodal task clearly differs from the text-only tasks considered earlier. 

% Our results demonstrate that greater improvement can be achieved on story videos by incorporating causal context. In particular, we discover that using a combination of causal and temporal context is better than pure temporal context in all settings. This highlights the importance of event causality in story understanding and opens the door for future research on incorporating event causality into story understanding tasks. 

% Research shows that temporal context may improve video understanding \cite{wu2019long,fan2019heterogeneous}. 

%Furthermore, the improvements achieved with the event graphs generated by ChatGPT attest to the correctness of the LLM-generated event graphs. 

\begin{table}[t!]
\centering
\resizebox{\columnwidth}{!}{
\begin{tabular}{@{}lll@{}}
\toprule
                     & Clip Acc. & Sent. IoU     \\
\midrule
\multicolumn{3}{c}{\emph{NeuMATCH Split (sub-sentence level)}} \\
NeuMATCH-MD (Supervised) & 4.0  & 2.4           \\
NeuMATCH-DTW (Supervised) & 10.3  & 7.5           \\
%NeuMATCH (Supervised) & 12.0 & 10.4 \\
SyMoN-MD & 5.9  &2.7 \\
%Temporal Context-MD & 10.8  & 5.7          \\
Temporal Context-DTW & 12.3  & 7.1          \\
%Causal+Temporal Context-MD & 15.9 & 13.1 \\ 
Causal+Temporal Context-DTW & \textbf{23.2 ($\uparrow $10.9)}  & \textbf{18.4 ($\uparrow 10.9$)}          \\
\midrule
\multicolumn{3}{c}{\emph{SyMoN Split (sub-sentence level)}}\\
SyMoN-MD & 10.1  &1.9 \\
%Temporal Context-MD & 16.0  & 5.4          \\
Temporal Context-DTW & 10.2  & 8.0         \\
%Causal+Temporal Context-MD & 21.2 & 17.9 \\ 
Causal+Temporal Context-DTW & \textbf{24.2 ($\uparrow 8.2$)}  & \textbf{21.5 ($\uparrow $13.5)}          \\
\midrule
\multicolumn{3}{c}{\emph{NeuMATCH Split (sentence level)}} \\
SyMoN-MD & 7.4  &3.4 \\
%Temporal Context-MD & 14.9  & 6.8          \\
Temporal Context-DTW & 29.2  & 18.3        \\
%Causal+Temporal Context-MD & 11.4 & 7.2 \\ 
Causal+Temporal Context-DTW & \textbf{33.3 ($\uparrow 4.1$)}  & \textbf{22.5 ($\uparrow 4.2$)}          \\
\midrule
\multicolumn{3}{c}{\emph{SyMoN Split (sentence level)}}\\
SyMoN-MD & 7.7  &3.3 \\
%Temporal Context-MD & 16.6  & 8.3          \\
Temporal Context-DTW & 32.5  & 19.6        \\
%Causal+Temporal Context-MD & 26.0 & 17.9 \\ 
Causal+Temporal Context-DTW & \textbf{40.2 ($\uparrow 7.7$)}  & \textbf{27.6 ($\uparrow 8.0$)}          \\
\bottomrule
\end{tabular}
}
\caption{Alignment performance on YMS. The improvement over baseline is shown in the parentheses. The best performance in each section and the best improvements overall are in bold. }
\vspace{-0.2cm}
\label{tab:alignment_result}
\end{table}

\section{Conclusion}
In this paper, we propose a simple and effective in-context-learning method for extracting event causality from stories with LLMs. We match and outperform supervised baselines in event causality extraction. Furthermore, we validate the quality of the extracted event causality by applying them in downstream story understanding tasks. Experiments show that event causality assists story evaluation and video-text alignment, indicating the critical role of event causality in story understanding.
%Experiments show that event causality significantly improved story understanding performance. Interestingly, automatic story evaluation performance can be improved simply by add the word ``causal'' into the prompt.

\paragraph{Acknowledgments}
We gratefully acknowledge the support by the Nanyang Associate Professorship and the National Research Foundation Fellowship (NRF-NRFF13-2021-0006), Singapore. 
%Any opinions, findings, conclusions, or recommendations expressed in this material are those of the authors and do not reflect the views of the funding agencies.

% \section*{Acknowledgements}
% The author would like to acknowledge Alibaba-NTU joint research institute. This research is supported, in part, by the National Research Foundation Fellowship (No. NRF-NRFF13-2021-0006), Singapore, and NTU Start-Up Grant (No. 04INS000400C300). 

\section*{Limitations}
% ACL 2023 requires all submissions to have a section titled ``Limitations'', for discussing the limitations of the paper as a complement to the discussion of strengths in the main text. This section should occur after the conclusion, but before the references. It will not count towards the page limit.
% The discussion of limitations is mandatory. Papers without a limitation section will be desk-rejected without review.

% While we are open to different types of limitations, just mentioning that a set of results have been shown for English only probably does not reflect what we expect. 
% Mentioning that the method works mostly for languages with limited morphology, like English, is a much better alternative.
% In addition, limitations such as low scalability to long text, the requirement of large GPU resources, or other things that inspire crucial further investigation are welcome.
The scope of our research is limited to causality between events. As such, the results may not extend to other types of causality (e.g. causality between events and emotion, location or possession states). Additionally, our technique works best on stories with clear event boundaries. When the stories contain dialogues or when the events are unclear, the improvements achieved by the causal graphs are limited.  

Furthermore, our exploration of event causality is confined to fiction stories and does not involve other domains such as news and tweets. While stories are a reflection of real life, their distribution emphasizes drama over realism. 
%For example, in stories, people ``die'' significantly more than they ``breathe''.
Therefore, it is not immediately clear if the event causality could play similar roles in other domains.

%Additionally, our explorations are limited to English story text. While citing that the results are shown for English-only as a limitation is a cliche in NLP research, in story understanding the limitations extend beyond language. Different cultures operate on different principles and social norms, and using English stories means that our interpretation of event causality may be biased toward a westernized view. 

%\hl{could you plz check this paragraph}
Our research on event causality for automatic story evaluation is primarily focused on the overall score, while some other studies delve into scoring different dimensions of quality, such as coherence, logicality, and relevance \citep{chhun2022human, ke2022ctrleval,xie2023next}. We argue that event causality may contribute to more than one dimension, since a clear and accurate causal graph implies relevance among events, logical causality, and overall coherence. Exploring how event causality contributes to the quality dimensions could be an interesting line of future research. 

% These studies provide varying definitions for dimensions such as  and more. However, we believe that event causality cannot be simply categorized into specific dimensions but should be regarded as a combination of these dimensions. After all,  Still, 

\section*{Broader Impact}
% Scientific work published at ACL 2023 must comply with the ACL Ethics Policy.\footnote{\url{https://www.aclweb.org/portal/content/acl-code-ethics}} We encourage all authors to include an explicit ethics statement on the broader impact of the work, or other ethical considerations after the conclusion but before the references. The ethics statement will not count toward the page limit (8 pages for long, 4 pages for short papers).

In this paper, we explore the use of LLM-extracted event causality in story understanding. We recognize LLMs may inadvertently contain  bias derived from training data. Furthermore, the story content we use may contain the biases of their creators, as well as social biases from the time periods of production.

Consequently, the causal relationships generated in our study are not intended as unbiased presentations of social norms. For this reason, we urge researchers to take caution when relying on LLMs or stories as a source for learning cultural and social conventions.

%see if this is ok:
%This study primarily assesses ChatGPT's ability to identify causal relationships in stories, a capability built upon commonsense knowledge acquired by the Large Language Model (LLM). We operate under the assumption that the knowledge acquired by LLM does not contain discriminatory or biased information and aligns with human core values. We further assume that LLM refrains from generating statements that exhibit discrimination or bias. Consequently, the causal relationships generated in our study are not intended to unfairly associate unrelated innocent groups with guilt or unjust causal relationships.

\bibliography{main}

\begin{thebibliography}{83}
\expandafter\ifx\csname natexlab\endcsname\relax\def\natexlab#1{#1}\fi

\bibitem[{Ammanabrolu et~al.(2021)Ammanabrolu, Cheung, Broniec, and Riedl}]{ammanabrolu2021automated}
Prithviraj Ammanabrolu, Wesley Cheung, William Broniec, and Mark~O Riedl. 2021.
\newblock Automated storytelling via causal, commonsense plot ordering.
\newblock In \emph{Proceedings of the AAAI Conference on Artificial Intelligence}, volume~35, pages 5859--5867.

\bibitem[{Andrus et~al.(2022)Andrus, Nasiri, Cui, Cullen, and Fulda}]{andrus2022enhanced}
Berkeley~R Andrus, Yeganeh Nasiri, Shilong Cui, Benjamin Cullen, and Nancy Fulda. 2022.
\newblock Enhanced story comprehension for large language models through dynamic document-based knowledge graphs.
\newblock In \emph{Proceedings of the AAAI Conference on Artificial Intelligence}, volume~36, pages 10436--10444.

\bibitem[{Anil et~al.(2023)Anil, Dai, Firat, Johnson, Lepikhin, Passos, Shakeri, Taropa, Bailey, Chen et~al.}]{anil2023palm}
Rohan Anil, Andrew~M Dai, Orhan Firat, Melvin Johnson, Dmitry Lepikhin, Alexandre Passos, Siamak Shakeri, Emanuel Taropa, Paige Bailey, Zhifeng Chen, et~al. 2023.
\newblock Palm 2 technical report.
\newblock \emph{arXiv preprint arXiv:2305.10403}.

\bibitem[{Bae and Young(2008)}]{bae2008use}
Byung-Chull Bae and R~Michael Young. 2008.
\newblock A use of flashback and foreshadowing for surprise arousal in narrative using a plan-based approach.
\newblock In \emph{Interactive Storytelling: First Joint International Conference on Interactive Digital Storytelling, ICIDS 2008 Erfurt, Germany, November 26-29, 2008 Proceedings 1}, pages 156--167. Springer.

\bibitem[{Benesty et~al.(2009)Benesty, Chen, Huang, and Cohen}]{benesty2009pearson}
Jacob Benesty, Jingdong Chen, Yiteng Huang, and Israel Cohen. 2009.
\newblock Pearson correlation coefficient.
\newblock In \emph{Noise reduction in speech processing}, pages 37--40. Springer.

\bibitem[{Brenner(2010)}]{brenner2010creating}
Michael Brenner. 2010.
\newblock Creating dynamic story plots with continual multiagent planning.
\newblock In \emph{Proceedings of the AAAI Conference on Artificial Intelligence}, volume~24, pages 1517--1522.

\bibitem[{Caselli et~al.(2021)Caselli, Hovy, Palmer, and Vossen}]{caselli2021computational}
Tommaso Caselli, Eduard Hovy, Martha Palmer, and Piek Vossen. 2021.
\newblock \emph{Computational Analysis of Storylines: Making Sense of Events}.
\newblock Cambridge University Press.

\bibitem[{Chhun et~al.(2022)Chhun, Colombo, Clavel, and Suchanek}]{chhun2022human}
Cyril Chhun, Pierre Colombo, Chlo{\'e} Clavel, and Fabian~M Suchanek. 2022.
\newblock Of human criteria and automatic metrics: A benchmark of the evaluation of story generation.
\newblock \emph{arXiv preprint arXiv:2208.11646}.

\bibitem[{Chiang and Lee(2023)}]{chiang2023can}
Cheng-Han Chiang and Hung-yi Lee. 2023.
\newblock Can large language models be an alternative to human evaluations?
\newblock In \emph{Proceedings of the 2023 Annual Meeting of the Association for Computational Linguistics}. Association for Computational Linguistics.

\bibitem[{Colon-Hernandez et~al.(2023)Colon-Hernandez, Lieberman, Xin, Yin, Breazeal, and Chin}]{colon2023adversarial}
Pedro Colon-Hernandez, Henry Lieberman, Yida Xin, Claire Yin, Cynthia Breazeal, and Peter Chin. 2023.
\newblock Adversarial transformer language models for contextual commonsense inference.
\newblock \emph{arXiv preprint arXiv:2302.05406}.

\bibitem[{Dogan et~al.(2018)Dogan, Li, Sigal, and Gross}]{dogan2018neural}
Pelin Dogan, Boyang Li, Leonid Sigal, and Markus Gross. 2018.
\newblock A neural multi-sequence alignment technique (neumatch).
\newblock In \emph{Proceedings of the IEEE Conference on Computer Vision and Pattern Recognition}, pages 8749--8758.

\bibitem[{Dong et~al.(2023)Dong, Martin, and Callison-Burch}]{dong2023corrpus}
Yijiang Dong, Lara Martin, and Chris Callison-Burch. 2023.
\newblock Corrpus: Code-based structured prompting for neurosymbolic story understanding.
\newblock In \emph{Findings of the Association for Computational Linguistics: ACL 2023}, pages 13152--13168.

\bibitem[{Du et~al.(2021)Du, Ding, Liu, and Qin}]{du2021learning}
Li~Du, Xiao Ding, Ting Liu, and Bing Qin. 2021.
\newblock Learning event graph knowledge for abductive reasoning.
\newblock In \emph{Proceedings of the 59th Annual Meeting of the Association for Computational Linguistics and the 11th International Joint Conference on Natural Language Processing (Volume 1: Long Papers)}, pages 5181--5190.

\bibitem[{Fan et~al.(2018)Fan, Lewis, and Dauphin}]{fan2018hierarchical}
Angela Fan, Mike Lewis, and Yann Dauphin. 2018.
\newblock \href {https://aclanthology.org/P18-1082} {Hierarchical neural story generation}.
\newblock In \emph{Proceedings of the 56th Annual Meeting of the Association for Computational Linguistics (Volume 1: Long Papers)}, Melbourne, Australia. Association for Computational Linguistics.

\bibitem[{Fletcher and Bloom(1988)}]{fletcher1988causal}
Charles~R Fletcher and Charles~P Bloom. 1988.
\newblock Causal reasoning in the comprehension of simple narrative texts.
\newblock \emph{Journal of Memory and language}, 27(3):235--244.

\bibitem[{Gansner et~al.(2015)Gansner, Koutsofios, and North}]{dot-language-2015}
Emden~R. Gansner, Eleftherios Koutsofios, and Stephen North. 2015.
\newblock \href {https://www.graphviz.org/pdf/dotguide.pdf} {Drawing graphs with dot}.
\newblock Technical report.

\bibitem[{Gao et~al.(2023)Gao, Ding, Qin, and Liu}]{gao2023chatgpt}
Jinglong Gao, Xiao Ding, Bing Qin, and Ting Liu. 2023.
\newblock Is chatgpt a good causal reasoner? a comprehensive evaluation.
\newblock \emph{arXiv preprint arXiv:2305.07375}.

\bibitem[{Gernsbacher(1997)}]{gernsbacher1997coherence}
Morton~Ann Gernsbacher. 1997.
\newblock Coherence cues mapping during comprehension.
\newblock In \emph{Processing interclausal relationships}, pages 3--21. Psychology Press.

\bibitem[{Ghazarian et~al.(2019)Ghazarian, Wei, Galstyan, and Peng}]{ghazarian2019better}
Sarik Ghazarian, Johnny Tian-Zheng Wei, Aram Galstyan, and Nanyun Peng. 2019.
\newblock Better automatic evaluation of open-domain dialogue systems with contextualized embeddings.
\newblock \emph{arXiv preprint arXiv:1904.10635}.

\bibitem[{Graesser et~al.(2003)Graesser, Olde, and Klettke}]{graesser200310}
Arthur~C Graesser, Brent Olde, and Bianca Klettke. 2003.
\newblock How does the mind construct and represent stories?
\newblock \emph{Narrative impact: Social and cognitive foundations}, page 121.

\bibitem[{Guan et~al.(2020)Guan, Huang, Zhao, Zhu, and Huang}]{guan2020knowledge}
Jian Guan, Fei Huang, Zhihao Zhao, Xiaoyan Zhu, and Minlie Huang. 2020.
\newblock A knowledge-enhanced pretraining model for commonsense story generation.
\newblock \emph{Transactions of the Association for Computational Linguistics}, 8:93--108.

\bibitem[{Guan and Huang(2020)}]{guan2020union}
Jian Guan and Minlie Huang. 2020.
\newblock Union: An unreferenced metric for evaluating open-ended story generation.
\newblock \emph{arXiv preprint arXiv:2009.07602}.

\bibitem[{Guan et~al.(2021)Guan, Zhang, Feng, Liu, Ding, Mao, Fan, and Huang}]{guan2021openmeva}
Jian Guan, Zhexin Zhang, Zhuoer Feng, Zitao Liu, Wenbiao Ding, Xiaoxi Mao, Changjie Fan, and Minlie Huang. 2021.
\newblock Openmeva: A benchmark for evaluating open-ended story generation metrics.
\newblock In \emph{Proceedings of the 2021 Annual Meeting of the Association for Computational Linguistics}. Association for Computational Linguistics.

\bibitem[{Gurulingappa et~al.(2012)Gurulingappa, Rajput, Roberts, Fluck, Hofmann-Apitius, and Toldo}]{gurulingappa2012development}
Harsha Gurulingappa, Abdul~Mateen Rajput, Angus Roberts, Juliane Fluck, Martin Hofmann-Apitius, and Luca Toldo. 2012.
\newblock Development of a benchmark corpus to support the automatic extraction of drug-related adverse effects from medical case reports.
\newblock \emph{Journal of biomedical informatics}, 45(5):885--892.

\bibitem[{Gutmann and Hyvarinen(2010)}]{gutmann2010noise}
Michael~U Gutmann and Aapo Hyvarinen. 2010.
\newblock Noise-contrastive estimation: A new estimation principle for unnormalized statistical models.
\newblock pages 297--304.

\bibitem[{Hong et~al.(2023)Hong, Sayeed, Mehra, Demberg, and Schiele}]{hong2023visual}
Xudong Hong, Asad Sayeed, Khushboo Mehra, Vera Demberg, and Bernt Schiele. 2023.
\newblock Visual writing prompts: Character-grounded story generation with curated image sequences.
\newblock \emph{Transactions of the Association for Computational Linguistics}, 11:565--581.

\bibitem[{Ke et~al.(2022)Ke, Zhou, Lin, Li, Zhou, Zhu, and Huang}]{ke2022ctrleval}
Pei Ke, Hao Zhou, Yankai Lin, Peng Li, Jie Zhou, Xiaoyan Zhu, and Minlie Huang. 2022.
\newblock Ctrleval: An unsupervised reference-free metric for evaluating controlled text generation.
\newblock \emph{arXiv preprint arXiv:2204.00862}.

\bibitem[{Keefe and McDaniel(1993)}]{keefe1993time}
Dennis~E Keefe and Mark~A McDaniel. 1993.
\newblock The time course and durability of predictive inferences.
\newblock \emph{Journal of memory and language}, 32(4):446--463.

\bibitem[{Kelly et~al.(2023)Kelly, Calderwood, Wardrip-Fruin, and Mateas}]{Kelly-AIIDE-2023}
Jack Kelly, Alex Calderwood, Noah Wardrip-Fruin, and Michael Mateas. 2023.
\newblock There and back again: Extracting formal domains for controllable neurosymbolic story authoring.
\newblock In \emph{AIIDE}.

\bibitem[{Kendall(1938)}]{kendall1938new}
Maurice~G Kendall. 1938.
\newblock A new measure of rank correlation.
\newblock \emph{Biometrika}, 30(1/2):81--93.

\bibitem[{Kuipers(1984)}]{kuipers1984commonsense}
Benjamin Kuipers. 1984.
\newblock Commonsense reasoning about causality: deriving behavior from structure.
\newblock \emph{Artificial intelligence}, 24(1-3):169--203.

\bibitem[{Lebowitz(1985)}]{lebowitz1985story}
Michael Lebowitz. 1985.
\newblock Story-telling as planning and learning.
\newblock \emph{Poetics}, 14(6):483--502.

\bibitem[{Li and Riedl(2010)}]{li2010offline}
Boyang Li and Mark Riedl. 2010.
\newblock An offline planning approach to game plotline adaptation.
\newblock In \emph{Proceedings of the AAAI Conference on Artificial Intelligence and Interactive Digital Entertainment}, volume~6, pages 45--50.

\bibitem[{Li et~al.(2022)Li, Martin, and Callison-Burch}]{li2022cis2}
Bryan Li, Lara~J Martin, and Chris Callison-Burch. 2022.
\newblock Cis2: A simplified commonsense inference evaluation for story prose.
\newblock In \emph{ACL Workshop}.

\bibitem[{Lincoln(1999)}]{bruce_lincoln_myth:1999}
Bruce Lincoln. 1999.
\newblock \emph{Theorizing myth: Narrative, ideology, and scholarship}.
\newblock University of Chicago Press.

\bibitem[{Liu et~al.(2023)Liu, Yin, Zhang, Feng, and Zhao}]{liu2023magic}
Xiao Liu, Da~Yin, Chen Zhang, Yansong Feng, and Dongyan Zhao. 2023.
\newblock The magic of if: Investigating causal reasoning abilities in large language models of code.
\newblock \emph{arXiv preprint arXiv:2305.19213}.

\bibitem[{Luo et~al.(2020)Luo, Ji, Shi, Huang, Duan, Li, Li, Bharti, and Zhou}]{luo2020univl}
Huaishao Luo, Lei Ji, Botian Shi, Haoyang Huang, Nan Duan, Tianrui Li, Jason Li, Taroon Bharti, and Ming Zhou. 2020.
\newblock Univl: A unified video and language pre-training model for multimodal understanding and generation.
\newblock \emph{arXiv preprint arXiv:2002.06353}.

\bibitem[{Meehan(1976)}]{Meehan-TaleSpin-1976}
J.~Meehan. 1976.
\newblock \emph{The Metanovel: Writing Stories by Computers}.
\newblock Ph.D. thesis, Yale University.

\bibitem[{Morabia(2007)}]{morabia2007epidemiologic}
Alfredo Morabia. 2007.
\newblock Epidemiologic interactions, complexity, and the lonesome death of max von pettenkofer.
\newblock \emph{American journal of epidemiology}, 166(11):1233--1238.

\bibitem[{Mostafazadeh et~al.(2020)Mostafazadeh, Kalyanpur, Moon, Buchanan, Berkowitz, Biran, and Chu-Carroll}]{Mostafazadeh2020GLUCOSEGA}
N.~Mostafazadeh, Aditya Kalyanpur, Lori Moon, David~W. Buchanan, Lauren Berkowitz, Or~Biran, and Jennifer Chu-Carroll. 2020.
\newblock \href {https://api.semanticscholar.org/CorpusID:221739295} {Glucose: Generalized and contextualized story explanations}.
\newblock In \emph{Conference on Empirical Methods in Natural Language Processing}.

\bibitem[{Mostafazadeh et~al.(2016)Mostafazadeh, Chambers, He, Parikh, Batra, Vanderwende, Kohli, and Allen}]{mostafazadeh2016corpus}
Nasrin Mostafazadeh, Nathanael Chambers, Xiaodong He, Devi Parikh, Dhruv Batra, Lucy Vanderwende, Pushmeet Kohli, and James Allen. 2016.
\newblock A corpus and evaluation framework for deeper understanding of commonsense stories.
\newblock \emph{arXiv preprinto arXiv:1604.01696}.

\bibitem[{Oatley(2008)}]{oatley2008}
Keith Oatley. 2008.
\newblock The mind's flight simulator.
\newblock \emph{The Psychologist}, 21:1030--1032.

\bibitem[{Oppenheimer and Susser(2007)}]{oppenheimer2007invited}
Gerald~M Oppenheimer and Ezra Susser. 2007.
\newblock Invited commentary: The context and challenge of von pettenkofer's contributions to epidemiology.
\newblock \emph{American journal of epidemiology}, 166(11):1239--1241.

\bibitem[{Ouyang et~al.(2022)Ouyang, Wu, Jiang, Almeida, Wainwright, Mishkin, Zhang, Agarwal, Slama, Ray et~al.}]{ouyang2022training}
Long Ouyang, Jeffrey Wu, Xu~Jiang, Diogo Almeida, Carroll Wainwright, Pamela Mishkin, Chong Zhang, Sandhini Agarwal, Katarina Slama, Alex Ray, et~al. 2022.
\newblock Training language models to follow instructions with human feedback.
\newblock \emph{Advances in Neural Information Processing Systems}, 35:27730--27744.

\bibitem[{Papineni et~al.(2002)Papineni, Roukos, Ward, and Zhu}]{papineni2002bleu}
Kishore Papineni, Salim Roukos, Todd Ward, and Wei-Jing Zhu. 2002.
\newblock Bleu: a method for automatic evaluation of machine translation.
\newblock In \emph{Proceedings of the 40th annual meeting of the Association for Computational Linguistics}, pages 311--318.

\bibitem[{Pearl(2018)}]{Pearl2018-PEATBO}
Judea Pearl. 2018.
\newblock \emph{The Book of Why: The New Science of Cause and Effect}.
\newblock Basic Books, New York.

\bibitem[{Porteous et~al.(2011)Porteous, Teutenberg, Charles, and Cavazza}]{porteous2011controlling}
Julie Porteous, Jonathan Teutenberg, Fred Charles, and Marc Cavazza. 2011.
\newblock Controlling narrative time in interactive storytelling.
\newblock In \emph{The 10th International Conference on Autonomous Agents and Multiagent Systems-Volume 2}, pages 449--456.

\bibitem[{Post(2018)}]{post2018call}
Matt Post. 2018.
\newblock A call for clarity in reporting bleu scores.
\newblock In \emph{Proceedings of the Third Conference on Machine Translation: Research Papers}. Association for Computational Linguistics.

\bibitem[{Radford et~al.(2019)Radford, Wu, Child, Luan, Amodei, Sutskever et~al.}]{radford2019language}
Alec Radford, Jeffrey Wu, Rewon Child, David Luan, Dario Amodei, Ilya Sutskever, et~al. 2019.
\newblock Language models are unsupervised multitask learners.
\newblock \emph{OpenAI blog}, 1(8):9.

\bibitem[{Reichenbach(1991)}]{reichenbach1991direction}
Hans Reichenbach. 1991.
\newblock \emph{The direction of time}, volume~65.
\newblock Univ of California Press.

\bibitem[{Reimers and Gurevych(2019{\natexlab{a}})}]{reimers2019sentence}
Nils Reimers and Iryna Gurevych. 2019{\natexlab{a}}.
\newblock \href {https://arxiv.org/abs/1908.10084} {Sentence-bert: Sentence embeddings using siamese bert-networks}.
\newblock In \emph{Proceedings of the 2019 Conference on Empirical Methods in Natural Language Processing}. Association for Computational Linguistics.

\bibitem[{Reimers and Gurevych(2019{\natexlab{b}})}]{reimers-2019-sentence-bert}
Nils Reimers and Iryna Gurevych. 2019{\natexlab{b}}.
\newblock \href {http://arxiv.org/abs/1908.10084} {Sentence-bert: Sentence embeddings using siamese bert-networks}.
\newblock In \emph{Proceedings of the 2019 Conference on Empirical Methods in Natural Language Processing}. Association for Computational Linguistics.

\bibitem[{Riedl and Young(2010)}]{riedl2010narrative}
Mark~O Riedl and Robert~Michael Young. 2010.
\newblock Narrative planning: Balancing plot and character.
\newblock \emph{Journal of Artificial Intelligence Research}, 39:217--268.

\bibitem[{Roemmele et~al.(2011)Roemmele, Bejan, and Gordon}]{roemmele2011choice}
Melissa Roemmele, Cosmin~Adrian Bejan, and Andrew~S Gordon. 2011.
\newblock Choice of plausible alternatives: An evaluation of commonsense causal reasoning.
\newblock In \emph{2011 AAAI Spring Symposium Series}.

\bibitem[{Sang et~al.(2022)Sang, Mou, Yu, Yao, Li, and Stanton}]{2022tvshowguess}
Yisi Sang, Xiangyang Mou, Mo~Yu, Shunyu Yao, Jing Li, and Jeffrey Stanton. 2022.
\newblock Tvshowguess: Character comprehension in stories as speaker guessing.
\newblock In \emph{NAACL}.

\bibitem[{Sellam et~al.(2020)Sellam, Das, and Parikh}]{sellam2020bleurt}
Thibault Sellam, Dipanjan Das, and Ankur~P Parikh. 2020.
\newblock Bleurt: Learning robust metrics for text generation.
\newblock \emph{arXiv preprint arXiv:2004.04696}.

\bibitem[{Shen et~al.(2023)Shen, Cheng, You, and Bing}]{shen2023large}
Chenhui Shen, Liying Cheng, Yang You, and Lidong Bing. 2023.
\newblock Are large language models good evaluators for abstractive summarization?
\newblock \emph{arXiv preprint arXiv:2305.13091}.

\bibitem[{Shoham(1990)}]{SHOHAM1990213}
Yoav Shoham. 1990.
\newblock \href {https://doi.org/https://doi.org/10.1016/0364-0213(90)90003-F} {Nonmonotonic reasoning and causation}.
\newblock \emph{Cognitive Science}, 14(2):213--252.

\bibitem[{Simon and Muise(2022)}]{TattleTale2022}
Nisha Simon and Christian Muise. 2022.
\newblock Tattletale: Storytelling with planning and large language models.
\newblock In \emph{ICAPS Workshop on Scheduling and Planning Applications workshop}.

\bibitem[{Soo et~al.(2016)Soo, Chen, and Lee}]{soo2016generate}
Von-Wun Soo, Tai-Hsun Chen, and Chi-Mou Lee. 2016.
\newblock Generate causal story plots by monte carlo tree search based on common sense ontology.
\newblock In \emph{2016 Joint 8th International Conference on Soft Computing and Intelligent Systems (SCIS) and 17th International Symposium on Advanced Intelligent Systems (ISIS)}, pages 610--615. IEEE.

\bibitem[{Spiliopoulou et~al.(2022)Spiliopoulou, Pagnoni, Bisk, and Hovy}]{spiliopoulou2022events}
Evangelia Spiliopoulou, Artidoro Pagnoni, Yonatan Bisk, and Eduard Hovy. 2022.
\newblock Events realm: Event reasoning of entity states via language models.
\newblock \emph{arXiv preprint arXiv:2211.05392}.

\bibitem[{Sun et~al.(2022)Sun, Chao, Ji, and Li}]{sun2022synopses}
Yidan Sun, Qin Chao, Yangfeng Ji, and Boyang Li. 2022.
\newblock Synopses of movie narratives: a video-language dataset for story understanding.
\newblock \emph{arXiv preprint arXiv:2203.05711}.

\bibitem[{Swartjes(2010)}]{swartjes2010whose}
Ivo Martinus~Theodorus Swartjes. 2010.
\newblock \emph{Whose story is it anyway? How improv informs agency and authorship of emergent narrative}.
\newblock Ph.D. thesis, University of Twente.

\bibitem[{Tamborrino et~al.(2020)Tamborrino, Pellicano, Pannier, Voitot, and Naudin}]{tamborrino2020pre}
Alexandre Tamborrino, Nicola Pellicano, Baptiste Pannier, Pascal Voitot, and Louise Naudin. 2020.
\newblock Pre-training is (almost) all you need: An application to commonsense reasoning.
\newblock \emph{arXiv preprint arXiv:2004.14074}.

\bibitem[{Touvron et~al.(2023)Touvron, Martin, Stone, Albert, Almahairi, Babaei, Bashlykov, Batra, Bhargava, Bhosale et~al.}]{touvron2023llama}
Hugo Touvron, Louis Martin, Kevin Stone, Peter Albert, Amjad Almahairi, Yasmine Babaei, Nikolay Bashlykov, Soumya Batra, Prajjwal Bhargava, Shruti Bhosale, et~al. 2023.
\newblock Llama 2: Open foundation and fine-tuned chat models.
\newblock \emph{arXiv preprint arXiv:2307.09288}.

\bibitem[{Trabasso and Van Den~Broek(1985)}]{trabasso1985causal}
Tom Trabasso and Paul Van Den~Broek. 1985.
\newblock Causal thinking and the representation of narrative events.
\newblock \emph{Journal of memory and language}, 24(5):612--630.

\bibitem[{Trabasso et~al.(1989)Trabasso, Van~den Broek, and Suh}]{trabasso1989logical}
Tom Trabasso, Paul Van~den Broek, and So~Young Suh. 1989.
\newblock Logical necessity and transitivity of causal relations in stories.
\newblock \emph{Discourse processes}, 12(1):1--25.

\bibitem[{Van~den Broek et~al.(1996)Van~den Broek, Lorch, and Thurlow}]{van1996children}
Paul Van~den Broek, Elizabeth~Pugzles Lorch, and Richard Thurlow. 1996.
\newblock Children's and adults' memory for television stories: The role of causal factors, story-grammar categories, and hierarchical level.
\newblock \emph{Child development}, 67(6):3010--3028.

\bibitem[{Wang et~al.(2023{\natexlab{a}})Wang, Xu, Liu, Meng, and Bai}]{wang2023contextualized}
Hecong Wang, Erqian Xu, Pinxin Liu, Zijian Meng, and Zhen Bai. 2023{\natexlab{a}}.
\newblock Contextualized multi-step commonsense reasoning through context extension.

\bibitem[{Wang et~al.(2023{\natexlab{b}})Wang, Liang, Meng, Shi, Li, Xu, Qu, and Zhou}]{wang2023chatgpt}
Jiaan Wang, Yunlong Liang, Fandong Meng, Haoxiang Shi, Zhixu Li, Jinan Xu, Jianfeng Qu, and Jie Zhou. 2023{\natexlab{b}}.
\newblock Is chatgpt a good nlg evaluator? a preliminary study.
\newblock \emph{arXiv preprint arXiv:2303.04048}.

\bibitem[{Wang et~al.(2023{\natexlab{c}})Wang, Do, Zhang, Zhang, Wang, Fang, Song, Wong, and See}]{wang-etal-2023-cola}
Zhaowei Wang, Quyet~V. Do, Hongming Zhang, Jiayao Zhang, Weiqi Wang, Tianqing Fang, Yangqiu Song, Ginny Wong, and Simon See. 2023{\natexlab{c}}.
\newblock \href {https://doi.org/10.18653/v1/2023.acl-long.288} {{COLA}: Contextualized commonsense causal reasoning from the causal inference perspective}.
\newblock In \emph{Proceedings of the 61st Annual Meeting of the Association for Computational Linguistics (Volume 1: Long Papers)}, pages 5253--5271, Toronto, Canada. Association for Computational Linguistics.

\bibitem[{Wei et~al.(2021)Wei, Bosma, Zhao, Guu, Yu, Lester, Du, Dai, and Le}]{wei2021finetuned}
Jason Wei, Maarten Bosma, Vincent~Y Zhao, Kelvin Guu, Adams~Wei Yu, Brian Lester, Nan Du, Andrew~M Dai, and Quoc~V Le. 2021.
\newblock Finetuned language models are zero-shot learners.
\newblock \emph{arXiv preprint arXiv:2109.01652}.

\bibitem[{Xie et~al.(2022)Xie, Hu, Li, Bi, Xing, and Peng}]{xie2022psychology}
Yuqiang Xie, Yue Hu, Yunpeng Li, Guanqun Bi, Luxi Xing, and Wei Peng. 2022.
\newblock Psychology-guided controllable story generation.
\newblock \emph{COLING}.

\bibitem[{Xie et~al.(2023)Xie, Cohn, and Lau}]{xie2023next}
Zhuohan Xie, Trevor Cohn, and Jey~Han Lau. 2023.
\newblock The next chapter: A study of large language models in storytelling.
\newblock In \emph{Proceedings of the 16th International Natural Language Generation Conference}, pages 323--351.

\bibitem[{Xu et~al.(2022)Xu, Wang, Yu, Ritchie, Yao, Wu, Zhang, Li, Bradford, Sun, Hoang, Sang, Hou, Ma, Yang, Peng, Yu, and Warschauer}]{xu2022fairytaleqa}
Ying Xu, Dakuo Wang, Mo~Yu, Daniel Ritchie, Bingsheng Yao, Tongshuang Wu, Zheng Zhang, Toby Jia-Jun Li, Nora Bradford, Branda Sun, Tran~Bao Hoang, Yisi Sang, Yufang Hou, Xiaojuan Ma, Diyi Yang, Nanyun Peng, Zhou Yu, and Mark Warschauer. 2022.
\newblock Fantastic questions and where to find them: Fairytale{QA} -- an authentic dataset for narrative comprehension.
\newblock In \emph{Proceedings of the 60th Annual Meeting of the Association for Computational Linguistics (Volume 1: Long Papers)}. Association for Computational Linguistics.

\bibitem[{Yang et~al.(2023)Yang, Klein, Peng, and Tian}]{yang-etal-2023-doc}
Kevin Yang, Dan Klein, Nanyun Peng, and Yuandong Tian. 2023.
\newblock \href {https://doi.org/10.18653/v1/2023.acl-long.190} {{DOC}: Improving long story coherence with detailed outline control}.
\newblock In \emph{Proceedings of the 61st Annual Meeting of the Association for Computational Linguistics (Volume 1: Long Papers)}, pages 3378--3465, Toronto, Canada. Association for Computational Linguistics.

\bibitem[{Yang et~al.(2022)Yang, Tian, Peng, and Klein}]{yang2022re3}
Kevin Yang, Yuandong Tian, Nanyun Peng, and Dan Klein. 2022.
\newblock Re3: Generating longer stories with recursive reprompting and revision.
\newblock In \emph{Conference on Empirical Methods in Natural Language Processing}.

\bibitem[{Ye et~al.(2022)Ye, Cui, Shi, and Riedl}]{ye2022neural}
Anbang Ye, Christopher Cui, Taiwei Shi, and Mark~O. Riedl. 2022.
\newblock Neural story planning.
\newblock \emph{arXiv Preprint 2212.08718}.

\bibitem[{Young et~al.(1994)Young, Pollack, and Moore}]{young1994decomposition}
R~Michael Young, Martha~E Pollack, and Johanna~D Moore. 1994.
\newblock Decomposition and causality in partial-order planning.
\newblock In \emph{AIPS}, pages 188--194.

\bibitem[{Yuan et~al.(2021)Yuan, Neubig, and Liu}]{yuan2021bartscore}
Weizhe Yuan, Graham Neubig, and Pengfei Liu. 2021.
\newblock Bartscore: Evaluating generated text as text generation.
\newblock \emph{Advances in Neural Information Processing Systems}, 34:27263--27277.

\bibitem[{Zar(2005)}]{zar2005spearman}
Jerrold~H Zar. 2005.
\newblock Spearman rank correlation.
\newblock \emph{Encyclopedia of Biostatistics}, 7.

\bibitem[{Zhang et~al.(2022)Zhang, Zhang, Su, and Roth}]{zhang2022rock}
Jiayao Zhang, Hongming Zhang, Weijie Su, and Dan Roth. 2022.
\newblock Rock: Causal inference principles for reasoning about commonsense causality.
\newblock In \emph{International Conference on Machine Learning}, pages 26750--26771. PMLR.

\bibitem[{Zhang et~al.(2019)Zhang, Kishore, Wu, Weinberger, and Artzi}]{zhang2019bertscore}
Tianyi Zhang, Varsha Kishore, Felix Wu, Kilian~Q Weinberger, and Yoav Artzi. 2019.
\newblock Bertscore: Evaluating text generation with bert.
\newblock \emph{arXiv preprint arXiv:1904.09675}.

\end{thebibliography}
\bibliographystyle{acl_natbib}

\appendix
\section{Causality Extraction on Glucose}
\label{sec:appendix}
\subsection{GLUCOSE Causality Dimensions}
\label{app:glucose_dims}
GLUCOSE \cite{Mostafazadeh2020GLUCOSEGA} divides causality within a story into ten dimensions. Here X represents the current event: 
\begin{enumerate}[itemsep=1mm]
\item Event that directly causes or enables X
\item Emotion or basic human drive that motivates X
\item Location state that enables X
\item Possession state that enables X
\item Other attributes enabling X
\item Event that X directly causes or enables
\item An emotion that is caused by X
\item A change in location that X results in
\item A change of possession that X results in
\item Other changes in property that X results in
\end{enumerate}
In the paper, we focus on dimensions 1 and 6 as they are about event causal relations. 

\subsection {GLUCOSE Finetuning Setup}
\label{app:finetune_glucose}
\noindent We finetuned two types of LMs, Decoder-only (GPT2) and Encoder-Decoder (T5), on 590K causal statements from GLUCOSE. These statements comprise 290K positive samples and 300K negative samples. The statements that explain the causal relationship between events, states or emotions are positive samples generated by AMT workers from the original GLUCOSE dataset. When there was no causal relationship between two events/states/emotions, there was no statement generated by AMT works. In this case, we generate a simple negative statement: "No, escaped." for such events/states/emotions.

\paragraph{GPT-2}
We finetuned gpt2-large on 4 NVIDIA A6000 GPUs with a learning rate of $3 \times 10^{-5}$ for 10 epochs. The batchsize is set to 64. The weight decay factor is $5 \times 10^{-4}$. 15\% of the input tokens are masked at random.  

\paragraph{T5}
We finetuned T5-large on 4 NVIDIA A6000 GPUs for 10 epochs, using a batch size of 32 on each GPU. The learning rate was $5 \times 10^{-4}$ under the cosine schedule with a warmup for the first 500 steps, and we adjusted the weight decay factor to $1 \times 10^{-2}$. No masked tokens were applied.

\subsection{The Causal Graph Generation Prompt}
In the paper, the causal graph generation prompt (Figure \ref{fig:prompt}) contains an instruction and a number of examples. In Section \ref{sec:expt_glucose}, the examples are always the same six in-domain, random selected stories from the GLUCOSE training set and the COPES validation set respectively.  In Section \ref{sec:story_eval}, the causal graph generation stage uses the same six stories from the GLUCOSE training set. 
In Section \ref{sec:story_alignment}, we use a single manually written example story.  

\subsection{Comparison of Prompts}
\label{app:glucose_prompts}
Aside from the prompt shown in Figure \ref{fig:prompt}, we design 11 additional prompts for event causality extraction. 
In this section, we present a comprehensive list of the 12 prompts we experimented with for causality extraction. 

\paragraph{Basic Prompt}
%\vspace*{0.5cm}
\par We show the basic prompt we use for causality extraction in Figure \ref{fig:basic_prompt}, all of the other prompts in this section are variations of this prompt.

\begin{figure}[ht!]
\begin{tcolorbox}[]
\small
\tt
\noindent Your job is to find all the causalities in a story. \\
\noindent You will be given a list of events in the story. An event can be caused by another event, an emotion, a possession state, a location state or some other property. Similarly, the effect of an event can be another event, an emotion, a possession state, a location state or some other property. For every event in the story, find all its causes and effects. For the description of events, you should write the event id in the parentheses after the description. For descrption of emotions, possession states, location states or other properties, write the type of the description in the parantheses after the description. \\
Example Input: \\
Event 0: When Dan goes to school in the morning, he has to take the bus. \\
Event 1: One day Dan was running late, and missed the bus to school. \\
Event 2: Dan called his friend Pete, and asked for a ride to school. \\
Example Output: \\
Dan's routine of taking the bus to school(Event 0) >Results in> Dan taking the bus to school(other property) \\
Dan takes the bus to school(other property) >Causes/Enables> Dan to miss the bus (Event 1). \\
Dan missing the bus (Event 1) >Causes/Enables> Dan to call his friend Pete for a ride (Event 2) \\
Input:\\
Event 0: <S1>\\
Event 1: <S2>\\
Event 2: <S3>\\
\\
Output:\\
\textcolor[RGB]{101,42,150}{\texttt{[Output from ChatGPT]}}
\end{tcolorbox}

\caption{The basic prompt for event causality }
\label{fig:basic_prompt}
\end{figure}

\paragraph{Separate Cause and Effect}
Dimensions 1 to 5 of GLUCOSE represent the causes of the event and Dimensions 6 to 10 represent the effects of the event, it seems natural to generate the causes and effects separately.

The prompt instructions remain unchanged, but when generating dimensions 1 to 5, the causal statements of dimensions 6 to 10 are removed from the example outputs.

\paragraph{Prompts Containing Definitions of Causality}
Note that in the above prompt, we do not include any definitions of causality. Next, we experiment with prompts that define causality in 4 different ways. Each prompt replaces the first line of the Basic Prompt with one definition below. 
\begin{itemize}
    \item Multifactorial: Your job is to find all the causalities in a story using the multifactorial definition of causality: A causes B when, in combination with other factors, it is a necessary or sufficient condition for the occurrence of event B. 
    \item Interventionist: Your job is to find all the causalities in a story using the interventionist definition of causality: A causes B when changing or intervening in the occurrence of A results in a corresponding change in the occurrence of B.
    \item Probabilistic: Your job is to find all the causalities in a story using the probabilistic definition of causality: A causes B when the likelihood or probability of B happening is significantly higher when A occurs compared to when A does not.
    \item Counterfactual: Your job is to find all the causalities in a story using the counterfactual definition of causality: A causes B if and only if when A does not happen, B will not happen.
\end{itemize}

\paragraph{ChatGPT Generated Instruction}
We input the examples into ChatGPT and asked ChatGPT to generate an instruction. To maximize reproducibility, we set the temperature to 0 when using the OpenAI API. We replace the instruction part of the prompt with instruction shown in Figure \ref{fig:chatgpt_instruction_prompt}.

\begin{figure}

\begin{tcolorbox}[breakable, enhanced]
\small
\tt
\noindent Identify and describe the causal relationships among the events in the narrative, highlighting how one event leads to or influences another. 
\end{tcolorbox}
\caption{The prompt instruction generated by ChatGPT}
\label{fig:chatgpt_instruction_prompt}
\end{figure}
\paragraph{Natural Language Form Output}
We keep the instructions and example input of the prompt unchanged, but change the format of the example output so that it is a grammatically correct sentence. See Figure \ref{fig:natual_language_format}.

\begin{figure}[t]
\begin{tcolorbox}[]
\small
\tt
Example Output:\\
Dan's routine of taking the bus to school(Event 0) results in Dan taking the bus to school(other property) \\
Dan takes the bus to school(other property) enables Dan to miss the bus (Event 1). \\
Dan missing the bus (Event 1) causes Dan to call his friend Pete for a ride (Event 2)
\end{tcolorbox}

\caption{Constrain the output format to a natural language.}
\label{fig:natual_language_format}
\end{figure}

\paragraph{ChatGPT Generated Format}
First, we remove the examples from the Basic Prompt. Then we use ChatGPT to perform zero-shot detection of causal relations between story events. We take the format created by ChatGPT (Figure {\ref{fig:chatgpt_format}}) and use that to format the few-shot examples. Finally, we add the re-formatted examples back into the basic prompt to get this new prompt. 

%We ask ChatGPT to generate free-form causality, and replace the output format in the prompt with the ChatGPT-generated format shown in Figure \ref{fig:chatgpt_format}:

\begin{figure}[ht]
\begin{tcolorbox}[enhanced]
\small
\tt
Output:\\
Original Event ID: 0 \\
Event: Dan goes to school in the morning, Dan takes the bus \\
Effect: Dan missed the bus to school \\
Original Event ID: 1 \\
Event: Dan was running late, and missed the bus to school \\
Cause: Dan takes the bus to school \\
Effect: Dan asks for a ride to school \\
Emotional Effect: Dan feels worried \\
Original Event ID: 2 \\
Event: Dan calls his friend Pete and asks for a ride to school \\
Motivation: Dan feels worried \\
Cause: Dan missed the bus to school \\
Effect: Pete gave Dan a ride to school \\
Emotional Effect: Dan feels relieved
\end{tcolorbox}

\caption{The output format generated by ChatGPT}
\label{fig:chatgpt_format}
\end{figure}

\paragraph{Curated Examples}
In every other prompt, the few-shot examples in the prompt are randomly chosen from the GLUCOSE training set. 

In this prompt, we select high-quality examples from the training set. In GLUCOSE, every pair of events is annotated by one human worker, who judges if there is a causal relation between them. Each human annotator also has a quality score between 1 and 3. In this prompt, we pick only example stories that are completely annotated by annotators with quality scores of 3. 

% human annotators annotate whether a causal relationship exists between every event pair in a story. Each event pair in the training set is annotated by 1 annotator. Furthermore, GLUCOSE authors checked the quality of the annotation and assigned a quality score between 1 and 3 to each annotator, where 1 means the annotator produces low-quality annotations and 3 means the annotator produces high-quality annotations. For this prompt, we pick stories from the GLUCOSE training set where all of the event pairs in the story are annotated by an annotator with a quality score of 3. 

%For the above prompts, we use 6 examples randomly chosen from GLUCOSE training set. Since GLUCOSE gives a quality score for each annotator, we pick examples annotated by high-quality annotators to create this prompt.

\paragraph{Causal Graph}
This is the main prompt of Figure \ref{fig:prompt} that we use throughout the paper. The LLM is asked to generate a list of edges between the nodes. 

\paragraph{Event Chain} We ask the LLM to describe how events are connected in causal chains. The LLM should generate a complete chain at a time, instead of a single causal relation at a time. 
Figure \ref{fig:event_chain} shows the prompt.

\begin{figure}[t]
\begin{tcolorbox}[]
\small
\tt
\noindent Here is a list of events from a story. Trace the domino effect of events in the story and explain how one event led to the next.\\
Example Input: \\
Event 0: When Dan goes to school in the morning, he has to take the bus. \\
Event 1: One day Dan was running late, and missed the bus to school. \\
Event 2: Dan called his friend Pete, and asked for a ride to school. \\
Example Output: \\
Chain 0: Event 0 -> Dan takes the bus to school(other property) -> Event 1 -> Event 2\\
Input:\\
Event 0: <S1>\\
Event 1: <S2>\\
Event 2: <S3>\\
\\
Output:\\
\textcolor[RGB]{101,42,150}{\texttt{[Output from ChatGPT]}}
\end{tcolorbox}
\caption{The prompt for event causality extraction in the form of event chains.}
\label{fig:event_chain}
\end{figure}

% \paragraph{Prompt Ensemble} Finally, we assemble the results from different prompts to achieve better results. Here, a simple voting scheme is adopted, a sentence is marked as having a causality of dimension $d$ only when the causality is detected with more than $n$ prompts, $n$ is a predefined threshold set to 4. 

\subsection{Evaluation Protocols}
Before event causality extraction, we first divide the story text into a sequence of sentences with the NLTK sent\_tokenizer: \url{https://www.nltk.org/api/nltk.tokenize.html}

For BLEU score, we calculate with the SacreBLEU implementation \cite{post2018call}, with equal weights up to 4-grams at corpus level on the three-reference test set.

We use the sentence-transformer implementation \cite{reimers-2019-sentence-bert} to calculate the BERTscore, using the average of token embeddings from the \texttt{bert-nli-mean-tokens} model without considering the ``idf'' weight of each token. The average of token embeddings is also used in calculating the BERT Similarity.

See Table \ref{tab:prompt_dim16} for results on GLUCOSE dimensions 1 and 6.
The results reported in Tables \ref{tab:copes}, \ref{tab:glucose_bleu_16}, and \ref{tab:prompt_dim16} are all from a single run.

%\subsection{Experiment with different prompts}
%\label{app:glucose_results_prompts}
%In Table \ref{tab:prompt_dim10}, we show the results on ChatGPT using different prompts, we show the results average over 10 dims. In Table \ref{tab:prompt_dim16} shows the results averaged over dimensions 1 and 6. 

% \begin{table}[ht]
% \centering

% \begin{adjustbox}{width=0.45\textwidth}
% \begin{tabular}{@{}lcccc@{}}
% \toprule

% & BLEU & BERTScore & \makecell[c]{BERT \\ Similar.}  & F1 \\
% \midrule
% Multifactorial          & 31.78  & 57.74  & 52.75  & 44.33  \\
% Interventionist         & 31.70  & 57.27  & 52.10  & 43.94  \\
% Probabilistic           & 29.83  & 56.97  & 51.70  & 43.37  \\
% Counterfactual          & 31.51  & 49.92  & 42.51  & 39.32 \\ 
% No Explicit Definition  & 28.04  & 57.12  & 51.98  & \textbf{46.60} \\
% Causal Chain             & 18.28  & 46.10  & 38.23  & 36.28  \\
% Causal Graph             & 8.10   & 46.96  & 40.39  & 39.75\\
% Natural Language Output & 24.23  & 52.67  & 46.12  & 40.38\\
% ChatGPT instruction     & 19.52  & 44.42  & 35.63  & 35.43\\
% ChatGPT format          & 15.98  & 51.85  & 45.86  & 41.7\\
% Curated Examples        & 24.68  & 57.22  & 52.50  & 40.14 \\
% Basic Prompt            & 20.51  & 50.32  & 43.74  & 38.42 \\
% Ensemble                & \textbf{33.06} & \textbf{63.15} & \textbf{60.46} & \textbf{50.70} \\
% \bottomrule
% \end{tabular}
% \end{adjustbox}
% \caption{Results of different prompts on GPT 3.5, averaged over 10 dimensions.}
% \label{tab:prompt_dim10}
% \end{table}

\begin{table}[t]
\centering

\begin{adjustbox}{width=0.45\textwidth}
\begin{tabular}{@{}lcccc@{}}
\toprule

& BLEU & BERTScore & \makecell[c]{BERT \\ Similar.}  & F1 \\
\midrule
Basic Prompt            & 30.20  & 65.96  & 65.61  & 54.11\\
Separate Cause and Effect  & 30.37  & 59.90  & 54.71  & 59.82 \\
Multifactorial          & 35.95  & 60.45  & 55.82  & 52.85\\
Interventionist         & 36.11  & 59.50  & 54.24  & 52.28\\
Probabilistic           & 32.13  & 59.18  & 53.79  & 50.90\\
Counterfactual          & \textbf{37.54}  & 54.98  & 48.89  & 49.12\\ 
ChatGPT Generated Instruction     & 30.12  & 55.31  & 50.73  & 50.96\\
Natural Language Form Output & 31.66  & 68.51 & 66.57  & 54.44\\
ChatGPT Generated Format          & 24.23  & 68.44  & 68.93  & 54.75\\
Curated Examples        & 33.97  & 69.74  & 69.40  & 55.34\\

Causal Graph (Ours)             & 21.2   & \textbf{75.33}  & \textbf{80.89}  & \textbf{60.75}\\
Event Chain             & 23.29  & 49.63  & 42.01  & 43.71\\

% Ensemble & 35.29 & 72.71 & 73.32 & 60.58 \\

\bottomrule
\end{tabular}
\end{adjustbox}
\caption{Results of different prompts on GPT 3.5, averaged over dimensions 1 and 6.}
\label{tab:prompt_dim16}
\end{table}

%\subsection{Separately Report Results of 10 Dimensions}
%\label{app:glucose_results_dims}

%Tables \ref{tab:glucose_bleu}, \ref{tab:glucose_bertscore},  \ref{tab:glucose_bert_sim}, \ref{tab:glucose_f1} show the BLEU score, BERTScore, BERT Similarity and F1 score of each of the GLUCOSE dimensions.

% \section{Event Causality in Story Understanding}

\section{Open-ended Generated Story Evaluation}
\label{app:story_eval}
\paragraph{Story Evaluation}
The prompt for scoring the story is first introduced by \citet{wang2023chatgpt}, and we insert a single word `causal' into it for exploratory experiments, shown in Figure \ref{fig:scoring_prompt_causal}.

\begin{figure}
\begin{tcolorbox}[]
\label{frame:causal_prompt}
\small
\tt
\noindent Score the following storyline given the beginning of the story with one to five stars.\\
Where one star means ``Nonsense'', \\
two stars mean ``The storyline has some connections with the beginning, but is not understandable'', \\
three stars mean ``The storyline has some \textcolor[RGB]{220,20,60}{\textbf{causal}}  connections with the beginning and is understandable'', \\
four stars mean ``The storyline is \textcolor[RGB]{220,20,60}{\textbf{causally}} consistent with the beginning and possibly involves a few grammar mistakes'', \\
and five stars mean ``Perfect storyline with \textcolor[RGB]{220,20,60}{\textbf{causal}}  connections and perfect grammar''.
\vspace{1ex}

\noindent The beginning of the story: <prompt>

\noindent Storyline: <generated story>

\noindent Stars: 
\end{tcolorbox}
\caption{The story scoring prompt from \citet{wang2023chatgpt} with the word ``causal'' inserted.}
\label{fig:scoring_prompt_causal}
\end{figure}

Figure \ref{fig:scoring_prompt_two_stage} demonstrates the prompt for the method ``ChatGPT-causal-graph" on WP under the few-shot cross-domain setting, as introduced in \S\ref{sec:story_eval}

\begin{figure}[t]
\begin{tcolorbox}[]
\small
\tt
% \raggedright\noindent ... (the same scoring prompt used by \citet{wang2023chatgpt})\\
Score the following storyline given the beginning of the story with one to five stars.\\
Where one star means "Nonsense",\\
two stars mean "The storyline has some connections with the beginning, but is not understandable",\\
three stars mean "The storyline has some causal connections with the beginning and is understandable",\\
four stars mean "The storyline is causally consistent with the beginning and possibly involves a few grammar mistakes",\\
and five stars mean "Perfect storyline with causal connections and perfect grammar".\\
\\
We also provide causal connections analyzed by experts, where each event is represented as a node, and the causal connections between these nodes are listed.\\

Here are two examples:\\
Example1:\\
The beginning of the story: i was sitting on the bench today. \\
Storyline: i heard my neighbor 's dogs barking. i looked at his dog and realized it was a Monday. the monday still runs through monday. i called my neighbor to let her know how i felt.\\
Event graph:\\
Edge 0: (Node 0 -> Node 1)\\
Edge 1: (Node 1 -> Node 2)\\
Edge 2: (Node 2 -> Node 3)\\
Edge 3: (Node 2 -> Node 4)\\
Stars: 1.4\\
\\
Example2:\\
The beginning of the story: [FEMALE] is an actress who just turned 19. \\
Storyline: she had a bad acting performance in an upcoming movie. she was disappointed.  then, she decided to try something different. she filmed herself playing in her very own version.\\
Event graph:\\
Edge 0: (Node 0 -> Node 1)\\
Edge 1: (Node 1 -> Node 2)\\
Edge 2: (Node 2 -> Node 3)\\
Edge 3: (Node 3 -> Node 4)\\
Stars: 5\\
(End of examples)\\
\\
The beginning of the story: <prompt>\\
Storyline: <generated story>\\
Event graph: \\
\textcolor[RGB]{101,42,150}{\texttt{[list the event graph generated with ChatGPT, here is just an example:]}}\\
Edge 0: (Node 0 -> Node 1)\\
Edge 1: (Node 1 -> Node 2)\\
Edge 2: (Node 2 -> Node 3)\\
Edge 3: (Node 3 -> Node 4)\\
Your score should reward stories with rich causal chains and penalize those that lack or have confusing causal chains.
\end{tcolorbox}
\caption{The story prompt we propose.}
\label{fig:scoring_prompt_two_stage}
\end{figure}

\paragraph{Correlation Computation}
We calculate average correlations at two aggregation levels: dataset level and writing prompt level.

Given a set of $N$ writing prompt sentences and $M$ generative language models. The story generated by the $m^{th}$ model using the $n^{th}$ writing prompt is denoted as $T_{n,m}$. The scoring for $T_{n,m}$ from ChatGPT or human workers are denoted as $S_{L}(T_{n,m})$ or $S_{human}(T_{n,m})$. 

\vspace{0.5ex}
\noindent\emph{Dataset Level}
\begin{equation}
\begin{aligned}
\small
\mathrm{Corr}_\text{d}=&\rho( [S_{L}(T_{1,1}), \dots, S_{L}(T_{M,N})] , \\
  & [S_{human}(T_{1,1}), \dots, S_{human}(T_{M,N})] )
\end{aligned}  
\end{equation}

\noindent\emph{Writing Prompt Level}
\vspace{-0.3ex}
\begin{equation}
\begin{aligned}
\small
\mathrm{Corr}_\text{p} =& \frac{1}{N} \sum^{N}_{n=1} (\rho( [S_{L}(T_{1,n}), \dots, S_{L}(T_{M,n})] ,\\
&[S_{human}(T_{1,n}), \dots, S_{human}(T_{M,n})] ))\\
\end{aligned}  
\end{equation}

\begin{figure}[t]
 \centering
 \includegraphics[width=\linewidth]{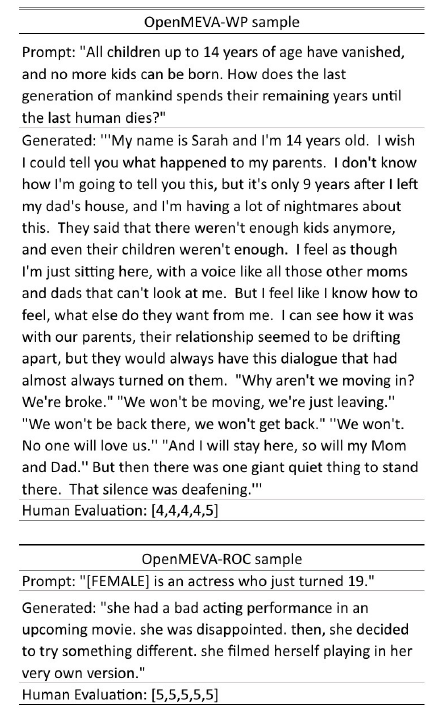}
 \caption{Example of OpenMEVA stories.}
\label{fig:open-meva-example}
\end{figure}

\section{Video-Text Alignment}
\label{app:alignment}
{\paragraph{Causal Graph Generation}}
The causal graph used in this experiment is generated with the prompt shown in Figure {\ref{fig:prompt}}. The LLM prompt includes an instruction and one example written by the authors. As the SyMoN stories are very long, we use a manually written story as demonstration instead of actual data samples from the SyMoN dataset. 

An average story in the SyMoN dataset contains 2408 tokens, however, the story length sometimes exceeds the context length of 4097 tokens. When this happens, we divide it into overlapping segments and concatenate the generated causal graphs. 

\paragraph{Training}
Following UniVL, we use S3D to extract 1 video feature per second. The video clips are trimmed or appended to 4 seconds. A video clip A is represented as 4 video feature vectors $\{v^{A}_1,\ldots,v^{A}_{4}\}$. The text chunks are trimmed or appended to 64 tokens. The video features and text tokens are then passed into the video and text encoders. Both encoders are 12-layer Transformers.

The model is trained on SyMoN with an initial learning rate of $5 \times 10^{-5}$ and cosine learning rate decay. We use a batch size of 256 and train for 40 epochs. The first epoch applies linear warm-up of learning rates. SGD with momentum of 0.9 is used for optimization and a weight decay term of 0.5 is added for 
regularization. 60\% of the text are masked at random. 

The number of parameters within the models is 153,784,064. The model is trained for 2.8 days on 1 NVIDIA A6000 1015 GPU. 

\paragraph{Evaluation}

In YMS, a text chunk may correspond to multiple video clips, whereas a video clip may correspond to one or zero text chunks. During inference, we first calculated the pair-wise similarity between every video clip and text chunk. Then we calculate the global alignment with DTW (introduced momentarily). If the similarity between an aligned video clip and text chunk falls below a threshold, tuned on the validation set, the video clip is considered to  not match anything.

The YMS dataset consists of 94 movie summary videos in total. In the NeuMatch split, the test set consists of 15 videos and the validation set consists of 12 videos. In the SyMoN split, the test set and validation set contain 65 and 29 videos respectively. 

The results in Table. \ref{tab:alignment_result} are from a single run.

\paragraph{Dynamic Time Wrapping}

DTW uses dynamic programming to find the best correspondence between two sequences based on distance (or similarity), the final alignment corresponds to the shortest distance or highest similarity. 

We use DTW to align a sequence of video clips $V=(v_1, \ldots, v_N)$ and a sequence of text chunks $T=(t_1, \ldots, t_M)$. We first assume that $v_1$ is aligned to $t_1$, thus the cost of matching $v_1$ and $t_1$ is $c(1,1)=0$, and the cost of match $v_1$ with $t_j(j \neq 1)$ is $c(1,j)=\infty$, and vice versa. Then, the minimal cost of aligning $(v_1, \ldots, v_i)$ with $(t_1, \ldots, t_j)$, can be calculated as:

\begin{equation}
\begin{split}
    c(i,j) = \min (c(i-1,j)+ d(i,j),\\c(i,j-1)+ d(i,j), \\ c(i-1,j-1)+ d(i,j))  
\end{split}
\end{equation}
where $d(i,j)$ denotes the distance between $v_i$ and $t_j$. Since we have the cosine similarity between each video-text pairs, the distance can be calculated as $d(i,j) = 1 - s(i,j)$, where $s(i,j)$ is the cosine similarity between $v_i$ and $t_j$.

\section{Case Studies of Identified Causal Relations}
\label{app:case_study}

In this section, we present case studies, including failure cases, of event causality identified using our approach.  Note that the \textcolor{teal}{[Correct]} / \textcolor{purple}{[Wrong]} labels are not present in the model output. We add them as part of our analysis.

\subsection{Case Study from COPES}
COPES contains annotations for causal predecessors of the last event only. The model output is correct on that edge. Edge 0 is a little ambiguous. One plausible interpretation is that the game motivates her to win. Another possible interpretation is that Alicia likes to win no matter what game she plays, so Edge 0 would be incorrect. We argue the relation in Edge 3 is correct because Alicia winning made her feel good, and the desire to repeat the good experience of winning motivated Alicia to play again.

\begin{figure}[t]
\begin{tcolorbox}[enhanced]
\small
\tt
Event List:\\
Node 0: Alicia was playing solitaire.\\
Node 1: She really wanted to win.\\
Node 2: Finally after twenty minutes, she won.\\
Node 3: Alicia decided to play again.\\
Node 4: This time she won in five minutes.\\
Output: \\
Edge 0: (Node 0 -> Node 1) [Ambiguous]\\
Edge 1: (Node 0 -> Node 2)\\
Edge 2: (Node 1 -> Node 2)\\
Edge 3: (Node 2 -> Node 3)\\
Edge 4: (Node 3 -> Node 4) \textcolor{teal}{[Correct]}\\
\end{tcolorbox}
\caption{Example of event graph for stories in the COPEs dataset.}
\label{fig:copes_example}
\end{figure}

\subsection{Case Study from OpenMEVA}
 OpenMEVA does not contain causal relation annotations. The labels reflect our own judgments. Edges 2 and 3 are ambiguous but it is possible that learning about vegan food allows my friend to teach other people. We consider Edge 4 wrong, as the friend teaching a class and my liking vegetarians seem unrelated. At the minimum, a number of additional events are needed to bridge the gap between the two statements.

\begin{figure}[t]
\begin{tcolorbox}[enhanced]
\small
\tt
Event List:\\
Node 0: My friend [FEMALE] became a vegan at age twenty.\\
Node 1: She bought many fruits and vegetables from the store.\\
Node 2: She learned a lot about vegan foods on the internet.\\
Node 3: She is now teaching a vegetarian class in her neighborhood.\\
Node 4: I am very happy to help out a vegetarian.\\
Output: \\
Edge 0: (Node 0 -> Node 1)\textcolor{teal}{[Correct]}\\
Edge 1: (Node 0 -> Node 2)\textcolor{teal}{[Correct]}\\
Edge 2: (Node 1 -> Node 3)[Ambiguous]\\
Edge 3: (Node 2 -> Node 3)[Ambiguous]\\
Edge 4: (Node 3 -> Node 4)\textcolor{purple}{[Wrong]}\\
\end{tcolorbox}
\caption{Example of event graph for stories in the OpenMEVA dataset.}
\label{fig:openmeva_example}
\end{figure}

\subsection{Case Study from SyMoN}
SyMoN does not contain causal relation annotations. The labels reflect our own judgments. Note that the LLM correctly singles out Node 2 as not causally related to any event. Edge 2 may appear dubious, as the action of asking for clothes does not immediately lead to getting dressed. However, one may reasonably infer that asking leads to receiving an answer, which enables getting dressed.

\begin{figure}[t]
\begin{tcolorbox}[enhanced]
\small
\tt
Event List:\\
Node 0: Tree wakes up in a dorm room that's not Tree's.\\
Node 1: Carter turns around and greets Tree.\\
Node 2. Tree's phone rings to the sound of a birthday song.\\
Node 3: Tree asks where Tree's clothes are.\\
Node 4: Tree immediately stands up to get dressed.\\
Node 5: Tree asks for Tylenol.\\
Node 6: Carter scrambles to find it.\\
…\\
Output: \\
Edge 0: (Node 0 -> Node 1) \textcolor{teal}{[Correct]}\\
Edge 1: (Node 0 -> Node 3) \textcolor{teal}{[Correct]}\\
Edge 2: (Node 3 -> Node 4) [Probable]\\
Edge 3: (Node 0 -> Node 5) \textcolor{teal}{[Correct]}\\
Edge 4: (Node 5 -> Node 6) \textcolor{teal}{[Correct]}\\
…\\
\end{tcolorbox}
\caption{Example of event graph for stories in the SyMoN dataset.}
\label{fig:symon_example}
\end{figure}

\section{Licensing Information}
The Glucose dataset is licensed under the Creative Commons Attribution-NonCommercial 4.0 International Public License. The COPES dataset is licensed under the MIT License. The OpenMEVA dataset is from \url{https://github.com/thu-coai/UNION}. The SyMoN dataset is from \url{https://github.com/insundaycathy/SYMON}. The YMS dataset is from \url{https://github.com/pelindogan/NeuMATCH/tree/master}. 

ChatGPT is under the GNU AFFERO GENERAL PUBLIC LICENSE Version 3. Llama-2 is licensed under LLAMA 2 COMMUNITY LICENSE AGREEMENT. Yi-34b-chat is licensed under  Yi Series Models Community License Agreement Version: 2.1. Falcon is licensed under Apache License Version 2.0. The Union model is from \url{https://github.com/thu-coai/UNION}. The UniVL model is licensed under the MIT License. 

\end{document}